\documentclass[11pt]{article}
\usepackage[preprint]{acl}
\usepackage{times}
\usepackage{latexsym}
\usepackage[T1]{fontenc}
\usepackage{iftex}\ifPDFTeX\usepackage[utf8]{inputenc}\fi
\usepackage{microtype}
\usepackage{amsmath,amssymb,mathtools}
\usepackage{graphicx,booktabs,array,longtable,tabularx,multirow}
\usepackage{enumitem}
\usepackage{caption,subcaption}
\usepackage{float,placeins}
\usepackage{listings}
\usepackage{url}
\usepackage{tikz}
\usepackage{algorithm}
\usepackage{algpseudocode}
\usetikzlibrary{arrows.meta,positioning,fit,backgrounds,calc}
\newcommand{\methodname}{Bayesian Model Discovery}
\newcommand{\vx}{\mathbf{x}}
\newcommand{\vtheta}{\boldsymbol{\theta}}

\newcommand{\nmsle}{\mathrm{nMSLE}}

\newcommand{\data}{\mathcal{D}}
\newcommand{\pool}{\mathcal{M}}

\definecolor{Gray}{gray}{0.9}
\definecolor{Red}{rgb}{0.768, 0.054, 0.054}
\definecolor{Blue}{rgb}{0.152, 0.294, 0.925}
\definecolor{URL}{rgb}{0,0.4,0.7}
\definecolor{Green}{rgb}{0,0.9,0}

\hypersetup{
    pdftitle={Bayesian Model Discovery with Language Models: Closing the Loop between Hypotheses and Experiments},
    pdfauthor={JooKyung Yuh, Byungmin Park, Eunseo Choi, Jinheon Baek, Junmo Cho, Kyu Hyun Choi, Sung Ju Hwang, Kyong Hwan Jin, Hae Beom Lee},
    colorlinks=true,
    citecolor=teal,
    linkcolor=Red,
    urlcolor=URL,
}

\setlist{nosep,leftmargin=*}
\title{Bayesian Model Discovery with Language Models:\\Closing the Loop between Hypotheses and Experiments}
\author{
  \textbf{JooKyung Yuh}$^{1}$ \quad
  \textbf{Byungmin Park}$^{1,*}$ \quad
  \textbf{Eunseo Choi}$^{1,*}$ \\
  \textbf{Jinheon Baek}$^{2}$ \quad
  \textbf{Junmo Cho}$^{2}$ \quad
  \textbf{Kyu Hyun Choi}$^{3}$ \\
  \textbf{Sung Ju Hwang}$^{2}$ \quad
  \textbf{Kyong Hwan Jin}$^{1,\dagger}$ \quad
  \textbf{Hae Beom Lee}$^{1,\dagger}$ \\[0.3em]
  \normalfont $^{1}$Korea University \quad $^{2}$KAIST \quad $^{3}$KETI \\[0.1em]
  \normalfont\small $^{*}$Equal contribution as second authors.\quad
  $^{\dagger}$Corresponding authors. \\[0.1em]
  \normalfont\small JooKyung Yuh: \texttt{johnlu0410@korea.ac.kr}\quad
  Hae Beom Lee: \texttt{haebeomlee@korea.ac.kr}
}
\date{}
\begin{document}
\maketitle
\begin{abstract}
Scientific model discovery requires proposing candidate models of an unknown system and designing experiments that discriminate among them. However, LLM-based approaches typically rely on in-context learning to predict outcomes and select experiments, lacking explicit model evaluation to systematically guide hypothesis search, while Bayesian approaches enable principled model comparison but are confined to a predefined candidate set which results in limited search spaces.
To this end, we introduce \methodname\ (BMD), which closes the loop between LLM-driven hypothesis generation and Bayesian experimental design.
Specifically, an LLM proposes model families, while a numerical engine fits their parameters, weighs them by model evidence, and averages their predictions.
Then, BMD chooses the next experimental conditions by maximizing the joint expected information gain over families and their parameters, while feeding the updated beliefs and residual diagnostics back to the LLM to guide its next proposals.
Across four scientific discovery benchmarks, BMD achieves lower final prediction error than in-context learning and LLM-based scientific agents while recovering exact functional forms more often, with ablations supporting numerical feedback and the joint objective.
\end{abstract}

\section{Introduction}
\label{sec:introduction}
Many scientific advances take the form of mathematical models that describe how a system responds to experimental conditions~\citep{langley1987scientific, schmidt2009distilling, wang2023scientific}.
Discovering such models alternates between proposing candidate models and designing experiments that discriminate among them~\citep{box1967discrimination, blei2014build}.
For example, a scientist studying how a drug slows down an enzyme may consider two candidate models: the drug either competes with the molecule the enzyme acts on or attaches elsewhere on the enzyme.
Both models may fit the initial measurements equally well, but the scientist can tell them apart with an experiment that adds a large amount of this molecule, as the two predict different reaction rates.
Recently, Large Language Models (LLMs) have shown promise in automating this process~\citep{boiko2023autonomous, romera2024funsearch, yang2024large, lu2024aiscientist, shojaee2025llmsr, baek2025researchagent}, owing to their broad scientific knowledge and reasoning abilities.

One common approach is \emph{in-context learning} (ICL)~\citep{brown2020gpt3}, where the LLM directly predicts responses and selects experiments from observations~\citep{gandhi2025boxinggym}.
While this approach exploits the prior knowledge of LLMs, it does not commit to explicit models fitted to data, leaving the LLM to make every prediction and choose experiments without quantifying which explanations remain plausible~\citep{gupta2025llms}.

In contrast, explicit probabilistic models address both issues by expressing each candidate explanation as an equation, called a \emph{model family}, whose functional form (e.g., competitive inhibition) contains unknown \emph{parameters} (e.g., the maximum reaction rate).
Specifically, once these parameters are estimated, the equation can predict responses at untested conditions, allowing a suitable family to generalize from few observations.
In addition, these models guide experiment selection, as Bayesian inference quantifies how plausible each family and its parameters remain~\citep{kass1995bayes, hoeting1999bma}, which Bayesian experimental design~\citep{lindley1956measure, chaloner1995bayesian, rainforth2024modern} uses to select the most informative experiment to run next.

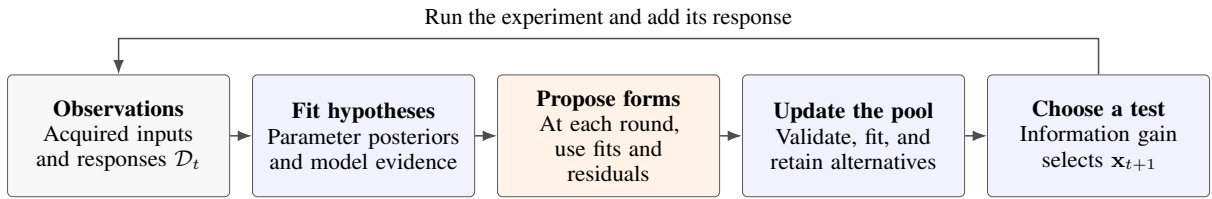
\begin{figure*}[t]
\centering
\resizebox{\textwidth}{!}{\begin{tikzpicture}[
  box/.style={draw=black!65,rounded corners=2pt,align=center,
    text width=27mm,minimum height=17mm,inner sep=2mm,font=\small},
  arrow/.style={-{Latex[length=2mm]},semithick,draw=black!70},
  note/.style={font=\footnotesize,align=center}
]
\node[box,fill=black!3] (data) at (0,0)
  {\textbf{Observations}\\Acquired inputs\\and responses $\data_t$};
\node[box,fill=blue!5] (fit) at (3.42,0)
  {\textbf{Fit hypotheses}\\Parameter posteriors\\and model evidence};
\node[box,fill=orange!10] (propose) at (6.84,0)
  {\textbf{Propose forms}\\At each round,\\use fits and residuals};
\node[box,fill=blue!5] (pool) at (10.26,0)
  {\textbf{Update the pool}\\Validate, fit, and\\retain alternatives};
\node[box,fill=blue!5] (design) at (13.68,0)
  {\textbf{Choose a test}\\Information gain\\selects $\vx_{t+1}$};
\draw[arrow] (data) -- (fit);
\draw[arrow] (fit) -- (propose);
\draw[arrow] (propose) -- (pool);
\draw[arrow] (pool) -- (design);
\draw[arrow] (design.north) -- ++(0,5mm) -|
  node[pos=0.25,above,font=\footnotesize] {Run the experiment and add its response} (data.north);
\end{tikzpicture}}
\caption{\textbf{Overview of \methodname.}
Given the observations $\data_t$, the numerical engine fits each candidate family and returns the resulting beliefs and residual diagnostics to the LLM, which proposes new families.
These proposals are validated, fitted, and compared with the existing families, after which the updated pool selects the next experimental conditions $\vx_{t+1}$ by maximizing the joint expected information gain.}
\label{fig:pipeline}
\vspace{-0.1in}
\end{figure*}

However, explicit modeling alone is not sufficient, as \citet{gandhi2025boxinggym} find no consistent benefit from augmenting an LLM with a statistical model.
Moreover, Bayesian inference over a fixed set of candidates cannot recover a family missing from it, while LLMs can propose new families~\citep{li2024model, shojaee2025llmsr} that remain unverified until they are fitted against the data.
The central challenge, therefore, lies in connecting family search, parameter inference, and experiment selection such that each step informs the others.

To address this, we present \methodname\ (BMD), a framework that closes the loop between LLM-driven hypothesis generation and Bayesian experimental design (Figure~\ref{fig:pipeline}).
Our key insight is that LLMs and Bayesian inference have complementary strengths: LLMs can draw on domain knowledge to propose plausible functional forms, whereas Bayesian inference can quantify how well each form explains the data.
Specifically, BMD employs an LLM to propose model families, while a numerical engine fits their parameters, weighs each family by its \emph{model evidence}, and averages their predictions.
Then, since a new observation can distinguish competing families, refine their parameters, or both, BMD chooses the next experimental conditions by maximizing the \emph{joint expected information gain} over families and their parameters~\citep{borth1975total, mcgree2017developments}, while feeding the updated beliefs and residual diagnostics back to the LLM to inform its next proposals.

We validate BMD on four scientific discovery benchmarks, namely ChemBench~\citep{kabra2026autoscilab}, BoxingGym~\citep{gandhi2025boxinggym}, ForceBench~\citep{wiemann2026discoverphysics}, and NewtonBench~\citep{zheng2026newtonbench}.
BMD achieves the lowest final prediction error on every benchmark against direct ICL and recent LLM-based scientific agents, while also recovering exact functional forms more often.
Our analyses further show that numerical feedback to the LLM improves prediction and that the joint objective outperforms targeting either families or parameters alone.
These findings suggest that LLMs are most effective as hypothesis generators within a closed loop of explicit model fitting and experimental design, rather than as standalone predictors of experimental results.

\section{Related Work}
\label{sec:related-work}

\paragraph{LLM-assisted model discovery.}
Symbolic regression searches for interpretable equations from
observations~\citep{schmidt2009distilling,brunton2016discovering,cranmer2023pysr}.
LLM-SR combines language-model proposals with numerical parameter
optimization~\citep{shojaee2025llmsr}, while BoxLM iterates
probabilistic-program generation, fitting, and criticism~\citep{li2024model}.
Other methods weight LLM-proposed models by their marginal
likelihood~\citep{domke2025large,wahl2026probabilistic}.
Box's Apprentice and LLM-AutoSciLab extend
model-based reasoning to interactive experimentation
~\citep{gandhi2025boxinggym,kabra2026autoscilab}.
Other closed-loop approaches select experiments by information gain over
LLM-proposed hypotheses, such as natural-language rules
~\citep{piriyakulkij2024doing} or cognitive models~\citep{prystawski2026autopsych}.
BMD searches over functional-form families using numerical parameter
posteriors and evidence-weighted predictions.

\paragraph{Bayesian experimental design.}
Bayesian experimental design selects observations to reduce
uncertainty~\citep{lindley1956measure,chaloner1995bayesian,rainforth2024modern}. Established criteria include
joint model discrimination and parameter estimation
~\citep{hill1968joint,borth1975total,mcgree2017developments}, and
expected predictive information gain (EPIG)~\citep{bickfordsmith2023epig}.
BMD uses joint EIG within an evolving pool of LLM-proposed families to
value both model discrimination and parameter refinement. We also
evaluate EPIG, which instead targets predictions at inputs of interest.
Recent work also applies EIG to LLM question asking, but estimates it
from the predictive distribution of the LLM itself rather than from fitted models
~\citep{choudhury2026bedllm}.

\paragraph{Concurrent work: Model Discovery Agent.}
Model Discovery Agent (MDA; \citealp{murphy2026mda}) similarly combines
LLM proposals, Bayesian inference, adaptive experimentation, and
numerical feedback. The cited MDA implementation uses a variance-based
surrogate for EIG, including a parameter-aware variant. BMD instead
estimates joint model--parameter EIG from fitted predictive likelihoods.
A second difference is the context supplied to the proposer. MDA
conditions proposals on domain context, whereas we anonymize variable
names and withhold task descriptions and semantic metadata on ChemBench,
BoxingGym, and NewtonBench. Numerical ranges and the modeling interface
remain available; ForceBench retains physical task metadata
(Section~\ref{sec:evaluation-protocol}). This evaluates discovery from
observations and feedback with fewer supplied domain cues.

\section{Approach}
\label{sec:approach}
\label{sec:method}
Our Bayesian Model Discovery (BMD) combines the flexibility of LLMs in proposing diverse functional forms with the strengths of Bayesian inference in evaluating hypotheses and quantifying uncertainty. By integrating these complementary capabilities, BMD iteratively discovers and refines scientific models through hypothesis generation, numerical evaluation, and experimental design (Figure~\ref{fig:pipeline};
Algorithm~\ref{alg:bmd}).

\subsection{Problem setup and overview}
\label{sec:problem}
\label{sec:model-families}
\label{sec:task}
\label{sec:endpoints}
An \emph{experimental design} $\vx\in\mathcal X$ specifies controllable
conditions, where
$\mathcal X$ is the set of allowed designs. The \emph{response} $y$
is the outcome measured under design $\vx$ and can be multi-dimensional.

A \emph{model family} $m$ specifies a functional form, such as a linear or saturating function. $\vtheta_m$ denotes a parameter vector associated with $m$. We assume that likelihood family is fixed for each task (e.g., Gaussian). $m$, $\vtheta_m$, and likelihood together define the conditional response distribution $p_m(y\mid\vx,\vtheta_m)$, expressed as a probability density or mass function. For each task, the true family $m^\star$ and its parameter vector $\vtheta^\star$ remain fixed and we can define
\[
    p^\star(y\mid\vx)
    :=p_{m^\star}(y\mid\vx,\vtheta^\star).
\]
The agent observes samples from this distribution but is not given
$m^\star$ or $\vtheta^\star$.

The initial dataset $\mathcal D_0$ contains $n_0$ observed
design--response pairs. With a budget of $B$ additional experiments,
at round $t=1,\ldots,B$, the agent chooses $\vx_t\in\mathcal X$ using
$\mathcal D_{t-1}$ and observes $y_t\sim p^\star(\cdot\mid\vx_t)$.
The available data are then
$\mathcal D_t=\mathcal D_0\cup\{(\vx_s,y_s)\}_{s=1}^t$.
The goals are accurate prediction at untested designs as well as recovery of
the underlying true functional form $m^\star$ and its parameter $\vtheta^\star$.

\subsection{Parameter inference and model evidence}
\label{sec:bmd-inference}
Unlike previous ICL-based approaches, our BMD separates the roles of the LLM and numerical inference. The LLM proposes candidate model families, while the numerical engine estimates the parameter posterior and model evidence for each candidate.

We first discuss posterior inference of model parameters within each candidate model family. We specify a prior $\pi_m(\vtheta_m)$ using
fixed coefficient-role rules.
Examples include $\mathcal N(0,1)$ for signed coefficients and
$\operatorname{LogNormal}(0,1)$ for positive coefficients.
The numerical engine assumes conditional independence of responses
given their designs and parameters, giving the likelihood
\begin{equation}
    L_{m,t}(\vtheta_m)
    =\prod_{(\vx,y)\in\mathcal D_t}
      p_m(y\mid\vx,\vtheta_m).
    \label{eq:bmd-likelihood}
\end{equation}
Then, Bayes' rule yields the parameter posterior and its normalizing
constant $Z_{m,t}= p(\mathcal D_t \mid m)$, called the \emph{model evidence}:
\begin{equation}
    \begin{aligned}
        p(\vtheta_m\mid\mathcal D_t,m)
        &=\frac{\pi_m(\vtheta_m)L_{m,t}(\vtheta_m)}{Z_{m,t}},\\
        Z_{m,t}
        &=\int \pi_m(\vtheta_m)L_{m,t}(\vtheta_m)
          \,\mathrm d\vtheta_m.
    \end{aligned}
    \label{eq:bmd-posterior-evidence}
\end{equation}
The posterior $p(\vtheta_m\mid\mathcal D_t,m)$ describes the uncertainty of $\vtheta_m$ within the family $m$, given the data $\mathcal D_t$. The evidence $Z_{m,t}$
measures its support by averaging the likelihood over the prior. Evidence thus
depends on the specified prior as well as the functional form.
We use sequential Monte Carlo (SMC)~\citep{delmoral2006smc} to obtain posterior samples and
an evidence estimate $\widehat Z_{m,t}$. Those posterior samples and evidence estimate are the core ingredients of BMD, which we explain in the next subsections. Appendix~\ref{app:method} gives the numerical settings.

\makeatletter
\newcommand{\BMDState}[1]{%
  \State \parbox[t]{\dimexpr\linewidth-\ALG@tlm\relax}{\strut#1\strut}%
}
\makeatother
\newlength{\BMDRequireExcess}
\newcommand{\BMDRequire}[1]{%
  \Require%
  \settowidth{\BMDRequireExcess}{\algorithmicrequire}%
  \addtolength{\BMDRequireExcess}{-\labelwidth}%
  \parbox[t]{\dimexpr\linewidth-\BMDRequireExcess\relax}{\strut#1\strut}%
}

\begin{algorithm}[t]
\caption{Bayesian Model Discovery (BMD)}
\label{alg:bmd}
\small
\begin{algorithmic}[1]
\BMDRequire{Initial data $\mathcal D_0$, design space $\mathcal X$,
            budget $B$, pool capacity $K$.}
\BMDState{Initialize candidate set $\mathcal C$ with LLM proposals from $\mathcal D_0$.}
\For{$t=0,\ldots,B$}
    \BMDState{\textbf{Fit hypotheses:} infer parameter posteriors and model evidence for $\mathcal C$ using $\mathcal D_t$.}
    \BMDState{\textbf{Propose forms:} for $t\geq1$, expand $\mathcal C$ with fit- and diagnostic-guided LLM proposals.}
    \BMDState{\textbf{Update the pool:} fit new proposals and retain up to $K$ highest-evidence families as $\pool_t$.}
    \BMDState{Form the predictive mixture $q_t$ over $\pool_t$.}
    \If{$t<B$}
        \BMDState{\textbf{Choose a test:} optimize $a_{t}$ over $\mathcal X$ to select $\vx_{t+1}$.}
        \BMDState{Observe $y_{t+1}$ and add $(\vx_{t+1},y_{t+1})$ to $\mathcal D_t$ to obtain $\mathcal D_{t+1}$.}
        \BMDState{Set $\mathcal C\gets\pool_t$.}
    \EndIf
\EndFor
\BMDState{\Return $q_B$, $\pool_B$, and its highest-evidence family.}
\end{algorithmic}
\end{algorithm}

\subsection{Model comparison and prediction}
\label{sec:bmd-prediction}
We next discuss posterior inference over candidate model families to assess which functional forms are more plausible given the observations. For the retained pool $\pool_t$, equal family-level prior weights give the posterior model probabilities
\begin{equation}
w_{m,t}
=\frac{\widehat Z_{m,t}}
{\sum_{j\in\pool_t}\widehat Z_{j,t}},
\qquad m\in\pool_t,
\label{eq}
\end{equation}
These probabilities are defined within the current fixed-sized model pool, rather than over all possible functional forms. Restricting inference to a finite pool makes model comparison tractable, although the true family need not belong to the pool. Section~\ref{sec:bmd-proposals} explains how the pool
evolves through LLM-generated proposals.

To predict while accounting for parameter uncertainty, each family
averages its likelihood over its posterior, giving the
\emph{posterior predictive distribution} $q_{m,t}$:
\begin{equation}
    \begin{aligned}
        q_{m,t}(y\mid\vx)
        &=\int p_m(y\mid\vx,\vtheta_m)\\[-2pt]
        &\qquad\times p(\vtheta_m\mid\mathcal D_t,m)
          \,\mathrm d\vtheta_m.
    \end{aligned}
    \label{eq:bmd-family-predictive}
\end{equation}
Bayesian model averaging (BMA) also accounts for uncertainty about the
family~\citep{hoeting1999bma}:
\begin{equation}
    q_t(y\mid\vx)
    =\sum_{m\in\pool_t}w_{m,t}\,q_{m,t}(y\mid\vx).
    \label{eq:bmd-predictive-mixture}
\end{equation}
For mean-response targets, BMD predicts with the mean of $q_t$.
For structural evaluation, it returns the highest-evidence family.

\subsection{Information-based experimental design}
\label{sec:bmd-design}
BMD selects experiments using joint expected information gain (EIG),
which targets uncertainty in both the model family and its parameters.

\paragraph{Joint model--parameter EIG.}
For a candidate design $\vx$, let $M$ be a random family drawn from
$\pool_t$ with probabilities $w_{m,t}$. Conditional on $M=m$, draw
its parameter vector $\boldsymbol\Theta_M$ from
$p(\vtheta_m\mid\mathcal D_t,m)$, and then draw a hypothetical response
$Y_{\vx}$ from $p_M(\cdot\mid\vx,\boldsymbol\Theta_M)$.
Its marginal distribution is $q_t(\cdot\mid\vx)$.
Using $I$ for mutual information, joint EIG is
\begin{equation}
    \begin{aligned}
        a_{t}(\vx)
        &=I(Y_{\vx};M,\boldsymbol\Theta_M\mid\mathcal D_t)\\
        &=\mathbb E\!\left[
            \log\frac{p_M(Y_{\vx}\mid\vx,\boldsymbol\Theta_M)}
                      {q_t(Y_{\vx}\mid\vx)}
          \right],
    \end{aligned}
    \label{eq:bmd-joint-eig}
\end{equation}
where the expectation is over these draws. The numerator evaluates
an outcome under one family and parameter setting; the denominator
averages over the alternatives. The expected log ratio, $a_t(\vx)$,
thus measures how informative an observation at $\vx$ would be about
the model family and its parameters.

The mutual-information chain rule separates the two information sources:
\begin{equation}
    \begin{aligned}
        a_{t}(\vx)
        &=\underbrace{I(Y_{\vx};M\mid\mathcal D_t)}_
            {\text{information about the family}}\\
        &\quad+\underbrace{
            I(Y_{\vx};\boldsymbol\Theta_M\mid M,\mathcal D_t)}_
            {\text{information about its parameters}}.
    \end{aligned}
    \label{eq:bmd-joint-decomposition}
\end{equation}
The first term favors designs that distinguish families, such as the
linear and saturating forms above. The second favors designs that
reduce parameter uncertainty within each family, weighted by its
posterior model probability $w_{m,t}$.
We estimate joint EIG $a_t(\vx)$ by Monte Carlo averaging
the log likelihood ratios, using SMC posterior samples to
approximate the evidence-weighted predictive distribution
in the denominator. See Appendix~\ref{app:analysis-joint} for estimating $a_t$.

\paragraph{Design optimization.}
For each round, we seek
\begin{equation}
    \vx_{t+1}\in\operatorname*{arg\,max}_{\vx\in\mathcal X}
    a_{t}(\vx).
    \label{eq:bmd-next-design}
\end{equation}
We optimize the estimated acquisition $a_t$ over the feasible design space,
keeping the fitted pool fixed throughout each search. For instance, we perform gradient ascent from five random initializations and pick the best one. See Appendix~\ref{app:method} for more details about solving Eq.~\eqref{eq:bmd-next-design}. See Algorithm~\ref{alg:bmd} for the overall procedure of BMD.

\begin{table*}[t]
\centering
\small
\setlength{\tabcolsep}{3pt}
\renewcommand{\arraystretch}{1.08}
\begin{tabularx}{\textwidth}{
@{}l
>{\raggedright\arraybackslash\hsize=.95\hsize}X
>{\raggedright\arraybackslash\hsize=.95\hsize}X
>{\raggedright\arraybackslash\hsize=1.10\hsize}X@{}}
\toprule
Benchmark & Scientific setting
& Design $\rightarrow$ observation & Evaluation target \\
\midrule
ChemBench
& Enzyme kinetics
& Reaction conditions $\rightarrow$ rate
& Noise-free observation mean (nMSLE) \\
BoxingGym
& Ecological/statistical processes
& Task inputs $\rightarrow$ noisy responses
& Noise-free response mean (nMSE) \\
ForceBench
& Force and motion laws
& Conditions $\rightarrow$ trajectories
& Shifted-design trajectories (nMSE) \\
NewtonBench
& Counterfactual physical laws
& Settings $\rightarrow$ physical responses
& Noise-free governing law (nMSLE) \\
\bottomrule
\end{tabularx}
\vspace{-0.06in}
\caption{Overview of the four scientific discovery benchmarks.}
\label{tab:benchmarks}
\vspace{-0.06in}
\end{table*}

\subsection{LLM-guided model pool expansion}
\label{sec:bmd-proposals}
\label{sec:bmd-pool}

The preceding sections describe parameter and model inference,
BMA prediction, and information-based experiment selection.
However, all these operations are defined within the current
model pool. If the true functional form is missing, updating
posteriors and collecting further observations do not by themselves
introduce it. Discovery therefore requires a mechanism for
proposing new functional forms.

Bayesian inference can be extended to search over symbolic expressions,
but such methods typically rely on a predefined grammar and can be
computationally expensive. BMD instead draws on an LLM's learned
mathematical knowledge to propose new forms without exhaustively
enumerating expressions. Numerical feedback is supplied in context
to guide these proposals toward what the current candidates fail
to explain.

At each round $t\geq1$, after acquiring a new observation,
the engine refits the families in $\pool_{t-1}$ on the
updated dataset $\mathcal D_t$.
The LLM then receives the accumulated observations,
candidate functional forms, and two types of numerical feedback:
\emph{belief information}, including model evidence,
posterior model probabilities, parameter posterior summaries,
and one-step predictive log scores; and
\emph{diagnostics}, including model-wise standardized residuals.
Belief information indicates which models the data support,
while diagnostics reveal where their predictions fall short.
For example, systematic curvature in a linear model's residuals
may motivate a nonlinear alternative.
The LLM uses this feedback to propose new functional forms.

The engine validates and deduplicates the proposals and fits them on
the same $\mathcal D_t$. It retains up to $K$ highest-evidence
families from the existing and newly proposed candidates to form
$\pool_t$, where $K$ is the pool capacity. Model probabilities and
BMA predictions are then recomputed, and the updated mixture guides
the next experiment. The resulting observation informs the next
round of numerical inference and LLM proposals, closing the loop
between hypothesis generation and experimentation.
Appendix~\ref{app:method} gives implementation details,
including periodic reconsideration of retired models;
Appendix~\ref{app:prompts} provides the prompts.

\section{Experiments}
\label{sec:experiments}\label{sec:results}
We first evaluate BMD's predictive accuracy and functional-form recovery,
then examine the source of performance improvements.

\paragraph{Benchmarks and baselines.}

We use four simulated scientific discovery benchmarks:
\textbf{ChemBench}~\citep{kabra2026autoscilab},
\textbf{BoxingGym}~\citep{gandhi2025boxinggym},
\textbf{ForceBench}~\citep{wiemann2026discoverphysics},
and \textbf{NewtonBench}~\citep{zheng2026newtonbench}.
Table~\ref{tab:benchmarks} summarizes these benchmarks.
For baselines, we first consider in-context learning
(\textbf{ICL}), which selects experiments and predicts
directly from observations without fitting explicit models.
Our primary baseline, \textbf{LLM-AutoSciLab}~\citep{kabra2026autoscilab},
maintains LLM-proposed hypotheses but fits them by regression
and selects experiments by prediction disagreement rather
than evidence and EIG.
We also compare with \textbf{Box's Apprentice}~\citep{gandhi2025boxinggym}
and \textbf{DP-Agent}~\citep{wiemann2026discoverphysics},
the reference agents of BoxingGym and ForceBench, respectively.
All methods use Claude Opus 5.5 for their LLM calls.

\begin{figure*}[t]
\centering
\includegraphics[width=\textwidth]{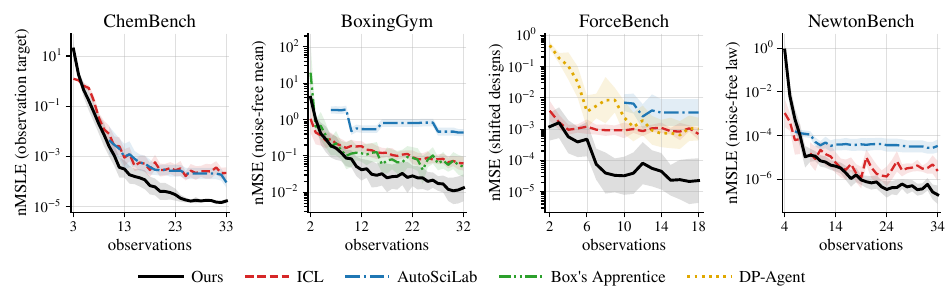}
\vspace{-0.35in}
\caption{\textbf{Predictive performance.}
Lower is better. Lines show geometric-mean prediction error,
aggregated over seeds within each task and then across tasks.
Shaded bands indicate seed dispersion in log space. The x-axis shows the total number of observations.
AutoSciLab curves start later because additional observations
are required before its first executable model is available.
Appendix~\ref{app:analysis-common} details the evaluation
protocol, aggregation, and exclusions.}
\label{fig:main-results-new}
\vspace{-0.15in}
\end{figure*}

\begin{figure*}[t]
\centering
\includegraphics[width=\textwidth]{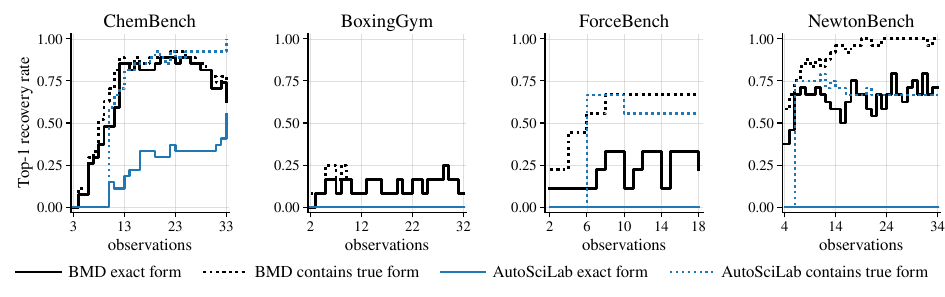}
\vspace{-0.35in}
\caption{\textbf{Strict recovery and ground-truth containment.}
Each curve shows the fraction of runs in which the selected top-1 model satisfies the corresponding criterion. \textbf{Strict recovery} (solid lines) requires structural equivalence to the ground-truth functional form. \textbf{Containment} (dotted lines) counts more general model families that can represent the ground truth as a special case. For example, a quadratic family contains a linear family when its quadratic coefficient is set to zero. Thus, containment does not necessarily imply recovery of the exact functional form. Appendix~\ref{app:analysis-discovery} details the evaluation criteria, including numerical checks.}
\label{fig:structural-recovery-main}
\end{figure*}

\paragraph{Information supplied to the agents.}
\label{sec:evaluation-protocol}
BMD and all baselines receive the same numerical ranges of the
design variables, with variable names anonymized.
Task descriptions and semantic metadata are withheld.
This eliminates explicit domain-specific semantic cues,
allowing us to better assess model discovery driven primarily
by observations and numerical feedback.
The exception is ForceBench, which is specifically designed
to use physical task metadata. We retain this information
for all methods to preserve the intended benchmark setting.
Appendix~\ref{app:analysis-common} details the information and
observation interfaces for each agent and benchmark.

\paragraph{Evaluation protocol.}
The main BMD runs use joint EIG. We report error against the total
observation count, using normalized mean squared logarithmic error
(nMSLE) or normalized mean squared error (nMSE) for the targets in
Table~\ref{tab:benchmarks}. Appendix~\ref{app:analysis-common}
specifies baseline schedules, missing-run exclusions, metrics, and
aggregation.

\subsection{Predictive performance and sample efficiency}
\label{sec:predictive-results}\label{sec:c1}
\label{sec:benchmarks}\label{sec:reporting}

We evaluate predictive performance and sample efficiency by tracking
prediction error as observations accumulate
(Figure~\ref{fig:main-results-new}).
Across all four benchmarks, BMD generally achieves lower error
at comparable observation budgets while also attaining the
lowest final error.
Overall, these results demonstrate BMD's superior predictive
performance and sample efficiency.

\raggedbottom

\begin{table*}[t]
\centering
\small
\setlength{\tabcolsep}{2pt}
\renewcommand{\arraystretch}{1.08}
\begin{tabular*}{\textwidth}{
@{\extracolsep{\fill}}lccc
@{\hspace{7pt}}ccccc
@{\hspace{7pt}}ccccc@{}}
\toprule
& \multicolumn{3}{c}{Components}
& \multicolumn{5}{c}{Relative error ($A=1$, $\downarrow$)}
& \multicolumn{5}{c}{GT identification (\%, $\uparrow$)}\\
\cmidrule(lr){2-4}\cmidrule(lr){5-9}\cmidrule(l){10-14}
Model & Explicit & SMC fit & Evidence
& Chem & Boxing & Force & Newton & Overall
& Chem & Boxing & Force & Newton & Overall\\
\midrule
A & \textsf{X} & \textsf{X} & \textsf{X}
& 1.000 & 1.000 & 1.000 & 1.000 & \textbf{1.000}
& -- & -- & -- & -- & --\\
B & \textsf{O} & \textsf{X} & \textsf{X}
& 0.917 & 0.868 & 0.230 & 0.777 & \textbf{0.613}
& 98.9 & 18.5 & 50.0 & 90.0 & \textbf{79.3}\\
C & \textsf{O} & \textsf{O} & \textsf{X}
& 0.351 & 1.388 & 0.174 & 0.219 & \textbf{0.369}
& 98.9 & 18.5 & 50.0 & 90.0 & \textbf{79.3}\\
D & \textsf{O} & \textsf{O} & \textsf{O}
& 0.349 & 0.667 & 0.036 & 0.247 & \textbf{0.214}
& 96.7 & 63.0 & 83.3 & 78.6 & \textbf{84.8}\\
\bottomrule
\end{tabular*}
\caption{\textbf{The role of numerical model evaluation.}
The methods progressively introduce (B) explicit model-based prediction
using an LLM-selected family and parameters, (C) Bayesian parameter
fitting via SMC for the family selected by B, and (D) evidence-based
selection instead of LLM selection.
Prediction errors are normalized by direct ICL (A) within each
benchmark; Overall is the geometric mean across benchmarks.}
\label{tab:explicit-summary}
\end{table*}

\subsection{Recovery of functional forms}
\label{sec:discovery-analysis}\label{sec:fitresults}\label{sec:c5}

Accurate prediction alone does not establish that an agent has
identified the underlying law. We therefore evaluate
\emph{strict recovery} (solid lines), which requires certified
structural equivalence to the true family, and
\emph{ground-truth containment} (dotted lines), which also accepts
a broader family capable of representing the true law.

At each observation budget, we evaluate BMD's highest-evidence
family and LLM-AutoSciLab's current hypothesis under both criteria.
Figure~\ref{fig:structural-recovery-main} reports the fraction of
runs that satisfy each criterion, denoted as Top-1 recovery rate.
Appendix~\ref{app:analysis-discovery} provides the evaluation details.

BMD demonstrates a clear advantage in exact-form discovery.
On BoxingGym, ForceBench, and NewtonBench, BMD successfully recovers
true functional forms, whereas AutoSciLab achieves no strict
recovery. On ChemBench, BMD identifies the correct structure much
earlier and maintains higher strict recovery at the final budget.
BMD also achieves higher containment on BoxingGym and NewtonBench.
The NewtonBench results are particularly revealing: AutoSciLab
frequently selects families that contain the ground truth but never
passes the strict criterion, whereas BMD achieves high recovery
under both criteria. These results demonstrate that BMD's advantage extends beyond predictive accuracy to structural recovery.

\begin{figure}[!t]
\centering
\includegraphics[width=\columnwidth]{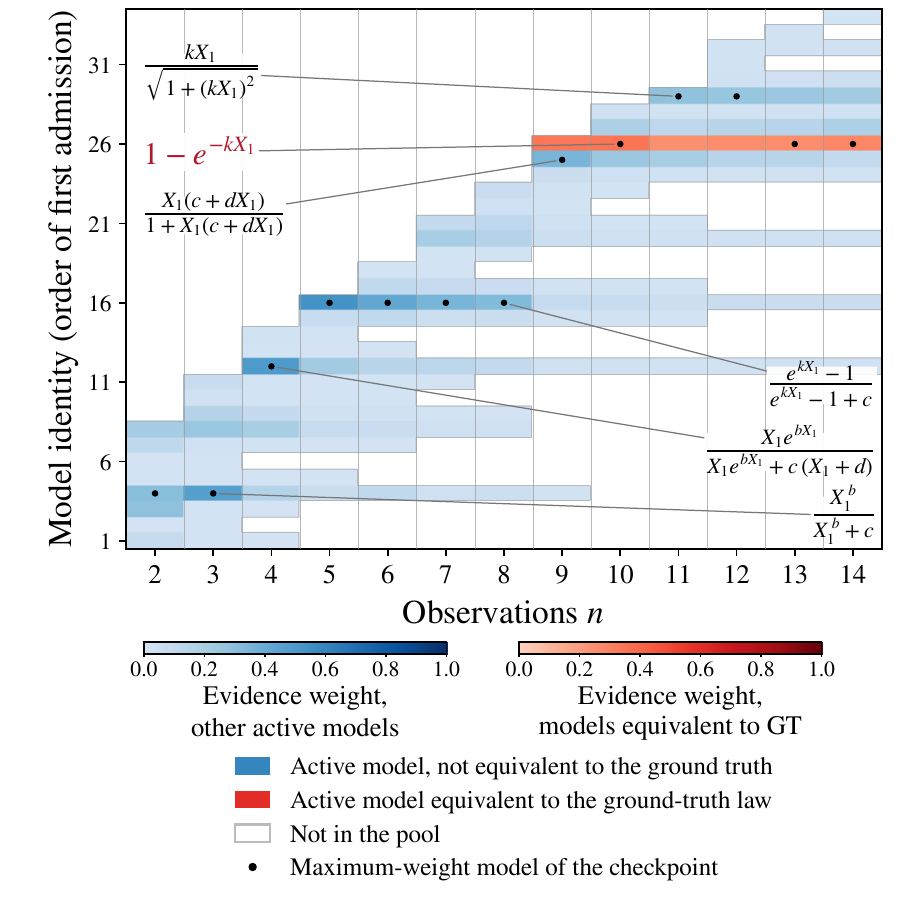}
\vspace{-0.3in}
\caption{\textbf{Exploring the model pool.} Rows follow first-admission order;
shading gives normalized evidence weights. Red marks ground-truth-equivalent
models, blue other active models, and white absence from the pool.
Black dots mark the highest-weight model. Annotations show candidate functional forms; \(X_1\) is the anonymized input, and letters denote coefficients.}
\label{fig:pool-evolution}
\end{figure}

\paragraph{A discovery trajectory.}
Figure~\ref{fig:pool-evolution} shows how the candidate pool evolves through experiments. The pool initially does not contain the ground-truth family, but at $n=9$ it introduces the ground-truth family
$1-e^{-kX_1}$.
This candidate becomes the highest-evidence model, remains in
the pool when another candidate temporarily leads, and later
returns to the top. This example illustrates how BMD introduces
missing functional forms and revises their support as observations
accumulate.

\begingroup
\captionsetup[figure]{hypcap=false}
\begin{figure*}[t]
\begin{minipage}[t]{0.48\textwidth}
\centering
\includegraphics[width=\linewidth]
{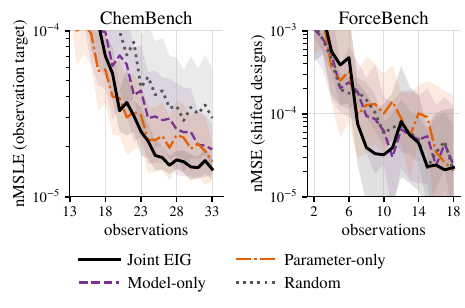}
\vspace{-0.35in}
\caption{\textbf{Joint EIG and its information components.}}
\label{fig:feedback}
\end{minipage}\hfill
\begin{minipage}[t]{0.48\textwidth}
\centering
\includegraphics[width=\linewidth]{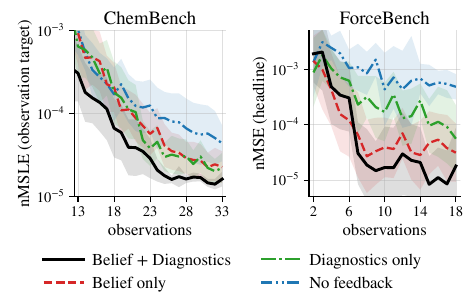}
\vspace{-0.35in}
\caption{\textbf{Belief and diagnostic feedback.}}
\label{fig:joint-analysis}
\end{minipage}
\end{figure*}
\endgroup

\subsection{The role of numerical model evaluation}
\label{sec:explicit-analysis}\label{sec:c2}

BMD uses the LLM to propose model families but delegates
parameter fitting and model comparison to a numerical engine.
Complementing the end-to-end results in Figure~\ref{fig:main-results-new}, we test
whether this division of labor improves upon the LLM's own
parameter estimates and model choices.

\paragraph{Experimental setup.}
For each world--seed pair, we collect 20 random-design
observations. Using only the first 10, Claude Opus 5.5
proposes 11 alternative families, which we combine with
the ground-truth (GT) family into a fixed pool of $K=12$
candidates. All methods share this pool and the full
observation set.

\paragraph{Models.} We consider the following methods.
\begin{itemize}[leftmargin=*, itemsep=1pt, topsep=2pt]
    \item \textbf{A (Direct ICL):}
    The LLM predicts responses directly from the observations
    and the pool.

    \item \textbf{B (LLM model selection + LLM parameters):}
The LLM selects a family from the pool and specifies
its parameter values based on the observations; the resulting
model is used for prediction.

    \item \textbf{C (LLM model selection + Bayesian parameter fitting):}
    Retains B's selected family but infers its parameter
    posterior using SMC and predicts with the posterior
    predictive mean.

    \item \textbf{D (Evidence-based model selection + Bayesian parameter fitting):}
Selects the highest-evidence family among SMC-fitted
candidates and predicts with its posterior predictive mean.
\end{itemize}
B--C compares parameter inference within the same family,
while C--D compares model selection under the same SMC
parameter-fitting procedure.
We report error on shared held-out points and GT identification
by structural equivalence of the selected family.

\paragraph{Results.}
Table~\ref{tab:explicit-summary} shows progressively lower
cross-benchmark geometric-mean error from A to B to C to D,
revealing three complementary benefits.

First, A--B shows that predicting through an explicit model
improves upon direct ICL, even when the LLM itself selects
the family and specifies its parameters.
Second, B--C shows that Bayesian parameter fitting further improves
prediction without changing the selected family,
highlighting the importance of numerical parameter inference.
Third, C--D shows that replacing LLM model selection with
evidence-based model selection improves aggregate prediction and
pooled top-1 GT identification (79.3\% to 84.8\%).

Overall, D reduces prediction error by a factor of 4.7
relative to A. These results show that predicting through
explicit models improves aggregate accuracy over direct ICL.
Once such models are introduced, however, selecting the
right family and fitting its parameters become critical.
Numerical parameter fitting and evidence-based selection
further improve performance over the LLM's own estimates
and choices. This motivates BMD's division of labor:
the LLM proposes functional forms, while the numerical
engine fits and compares them.
\subsection{The value of joint EIG}
\label{sec:joint-analysis}\label{sec:design-results}\label{sec:c4}

An informative experiment may distinguish competing model
families, refine their parameters, or both. Figure~\ref{fig:feedback} compares our Joint EIG acquisition function against Model-only EIG, Parameter-only EIG, and Random design (random selection of the next design point $\vx_{t+1}$).
Model-only EIG and Parameter-only EIG correspond to the terms in
Equation~\eqref{eq:bmd-joint-decomposition}.

Joint EIG achieves the lowest final prediction error on both benchmarks.
On ChemBench, it maintains a clear advantage over the alternatives
through much of the later observation budgets.
On ForceBench, Joint EIG reduces prediction error substantially faster
at intermediate budgets, although the other methods eventually
approach its performance.
These results highlight the benefit of jointly targeting model
discrimination and parameter refinement, suggesting that
accounting for both sources of uncertainty can lead to
more effective experimental designs.

\subsection{What feedback helps hypothesis discovery?}
\label{sec:feedback-analysis}\label{sec:c3}

Finally, beyond fitting and comparing candidate models, BMD feeds
numerical inference results back to the LLM to guide
subsequent hypothesis generation.
We examine whether this feedback improves discovery through
a $2\times2$ ablation: Full (belief + diagnostics),
Belief-only, Diagnostics-only, and No feedback.
Belief feedback includes model evidence, posterior weights,
and parameter summaries, while diagnostics provide
model-wise predictive residuals.
All variants share the same initialization and retain
Bayesian inference and joint-EIG experimental design.

Figure~\ref{fig:joint-analysis} shows that Full achieves the lowest final prediction
error on both ChemBench and ForceBench, while No feedback
performs worst. Either feedback type generally improves
prediction, although their relative contributions vary
across benchmarks.
These results suggest that numerical inference is valuable
not only for evaluating existing hypotheses but also for
guiding the LLM toward better subsequent proposals,
supporting BMD's closed-loop design.

\section{Conclusion}
\label{sec:conclusion}

BMD connects LLM hypothesis generation, Bayesian model
evaluation, and information-theoretic experimental design. Across four simulated benchmarks,
it improves prediction, sample efficiency, and exact-form
recovery. Controlled comparisons favor explicit modeling over
direct ICL, with further gains from Bayesian fitting and
evidence-based selection. Acquisition comparisons favor joint
model--parameter EIG, while feedback ablations show that
returning numerical beliefs and diagnostics to the LLM
improves downstream prediction, reinforcing BMD's closed-loop design.
\label{marker:main-content-end}
\section*{Limitations}
\label{sec:limitations}
Our experiments focus on simulated scientific tasks;
validation in real-world experimental settings remains
future work. Such evaluation would help assess robustness
beyond the observation models and experimental conditions
considered here. BMD's uncertainty estimates are conditional
on its current finite candidate pool rather than the space
of all possible laws. Although the pool evolves through
LLM proposals, discovery still depends on generating suitable
model families, and our structural-recovery evaluation
covers only targets expressible in the modeling interface.

\section*{Ethical Considerations}
Our experiments are conducted in simulated scientific
environments without human participants.
While BMD aims to assist scientific discovery, automatically
generated hypotheses may be inaccurate or misleading,
potentially leading to incorrect scientific conclusions.
Real-world applications should therefore involve independent
validation of discovered models and appropriate human oversight
before conducting consequential experiments.
\FloatBarrier
\label{marker:main-with-limitations-end}
\flushbottom
\bibliography{references}
\clearpage
\appendix
\section{Experimental Protocol and Metrics}
\label{app:analysis-protocol}\label{app:analysis-common}
\label{app:analysis-c1}\label{app:metrics}

\paragraph{Tasks, runs, and observation counts.}
A task specifies a generating mechanism; a world also fixes its
parameters and observation process. A run is one execution with a
particular seed. Table~\ref{tab:main-cohort} lists the task sets for the
sequential prediction and recovery experiments in
Sections~\ref{sec:c1} and~\ref{sec:c5}. The shared-candidate study in
Section~\ref{sec:c2} uses a separate 217-run cohort, specified in
Appendix~\ref{app:analysis-explicit}; its run counts should not be
substituted for those of the sequential experiments.
ChemBench, ForceBench, and NewtonBench keep each task's law and
constants fixed across seeds. BoxingGym instead draws parameters for
each task--seed world. Seeds control numerical randomness and, where randomized, initial
observations. The three standard seeds are 31, 37, and 41; BoxingGym uses
31, 41, and 47 for peregrines and 43, 53, and 61 for survival.
The peregrines substitution avoids degenerate initial data.

With $n_0$ initial observations and $t$ additional responses, the plotted
count is $n=n_0+t$. One ForceBench trajectory counts as one observation,
not one observation per time point. Repeated designs generate fresh
noise on BoxingGym and ForceBench, but return the same response on
ChemBench and NewtonBench, whose noise is a deterministic function of
the input. These simulator rules differ from the conditional-independence
assumption used for numerical fitting.

\begin{table*}[t]
\centering\small
\setlength{\tabcolsep}{4pt}
\caption{Main experimental tasks and budgets. Structural recovery uses
only targets representable in the interface; predictive evaluation
retains all tasks. Each task has three seeds.}
\label{tab:main-cohort}
\begin{tabularx}{\textwidth}{@{}p{0.16\textwidth}*{4}{>{\raggedright\arraybackslash}X}@{}}
\toprule
& ChemBench & BoxingGym & ForceBench & NewtonBench\\
\midrule
Tasks & Michaelis--Menten, product inhibition, ping-pong; each easy/medium/hard
& Death, dugongs, hyperbolic, peregrines, survival
& Gravity, Yukawa, Coulomb (easy), oscillator, extra dimensions
& Magnetic force, Fourier, Snell, radioactive decay, Malus, sound speed, Hooke, Bose--Einstein, heat transfer (medium)\\
BMD runs & 27 & 15 & 15 & 27\\
Recovery runs & 27 & 12 & 9 & 24\\
$n_0$ / additional / final $n$ & 3 / 30 / 33 & 2 / 30 / 32 & 2 / 16 / 18 & 4 / 30 / 34\\
Design space & 7 continuous reaction conditions
& Task-specific; hyperbolic has 3 integer inputs, survival a discrete patient menu
& 2 continuous parameters in $[0.1,10]^2$
& Task-specific continuous inputs\\
Prediction scored & After proposals & Before proposals & Before proposals & After proposals\\
\bottomrule
\end{tabularx}
\end{table*}

\paragraph{Information supplied to the agents.}
BMD and all baselines receive the same numerical ranges of the design
variables, with variable names anonymized on all four benchmarks.
Task descriptions and semantic metadata are withheld on ChemBench,
BoxingGym, and NewtonBench; BMD's modeling interface remains available.
ForceBench is the exception to metadata withholding: all methods retain
the same physical task metadata to preserve the intended benchmark
setting, while variable identifiers remain anonymized. There, the
interface identifies separation, elapsed time, and inertial mass, and
specifies how a force expression produces trajectories. For position
vector $\mathbf r$, the declared acceleration
is $-f_m/X_3$ along $\mathbf r/\|\mathbf r\|$, where $f_m$ is the proposed
expression and $X_3$ is the inertial-mass input. The initial position and
velocity and the velocity-Verlet grid ($\Delta t=0.005$) are supplied.
Each trajectory contains two coordinates at ten measurement times.
No world name, force law, or law catalogue is supplied to the proposer.
In BMD's discovery runs, all initial candidates are LLM proposals;
assigning a canonical catalogue name to a matching expression does not
insert that expression. The fixed-pool diagnostic in
Section~\ref{sec:c2} is the explicit exception: it inserts the GT family
to separate model evaluation from candidate-generation success.

\paragraph{Baseline interfaces and schedules.}
All methods use Claude Opus 5.5 for their LLM calls.
\emph{LLM-AutoSciLab} uses an every-round variant for predictive curves:
one experiment and one equation-learning step per round, with the same
LLM and reasoning effort as BMD. It follows the same anonymization and
metadata policy described above: in particular, it does not receive
native variable names or chemical task metadata on ChemBench.
Its NewtonBench recovery curve uses the released schedule, with the
first law at $n=16$.

The ForceBench comparisons use the rerun with a common information and
observation interface for BMD, direct ICL, LLM-AutoSciLab, and DP-Agent.
All receive the same physical task metadata and design constraints,
observe trajectories through the same interface, and are evaluated
using the same targets and scorer. Each trajectory counts as one
observation for every method. The agents may choose different designs
and hence observe different responses, but they do not receive different
types of experimental measurements.

\emph{Box's Apprentice} on BoxingGym and \emph{DP-Agent} on ForceBench
also use every-round variants. \emph{Direct ICL} chooses experimental
designs from its observation history and receives separate prediction
requests at each observation count; these requests do not change that
history. Different experiment-selection policies are part of the
comparison, whereas Section~\ref{sec:c2} holds the realized observations
and candidate families fixed across methods.

\paragraph{Prediction targets.}
Let $Q$ be the empirical distribution of evaluation inputs, $h(\vx)$
the evaluation target, and $\widehat\mu_n(\vx)$ the prediction after
$n$ observations. For nonnegative targets,
\begin{equation}
\nmsle_Q=
\frac{\mathbb E_Q[(\log(1+\widehat\mu_n)-\log(1+h))^2]}
{\operatorname{Var}_Q[\log(1+h)]}.
\label{eq:appendix-nmsle}
\end{equation}
The nMSE divides the mean squared error by the benchmark's fixed
variance normalizer, averaging over the prescribed coordinates and
time points for trajectories. Normalizers are specific to the target
and evaluation set.

ChemBench evaluates the noise-free observation mean
$g^\star(\vx)=f^\star(\vx)s(T,\mathrm{pH})$, including environmental
effects $s$, rather than the primary kinetic law $f^\star$ alone.
BoxingGym uses noise-free response-mean nMSE, including survival.
NewtonBench evaluates its noise-free governing law by nMSLE.
ChemBench and NewtonBench use 256 fixed evaluation inputs per task,
shared across methods and seeds. BoxingGym draws 256 inputs per world
and excludes observed designs. ForceBench evaluates trajectory nMSE on
six fixed parameter settings,
$p_1\in\{3,4,5\}$ and $p_2\in\{3,5\}$, inside the acquisition box.
These settings are our evaluation set, not the benchmark's three
standard scoring designs; the normalizer is the benchmark's per-world
variance constant, with a fixed scoring seed.

The targets matter: low observation-mean error need not imply low
primary-law error. For $u=\log(1+f^\star)$,
$v=\log(1+g^\star)$, and $\widehat v=\log(1+\widehat\mu_n)$,
\begin{equation}
\begin{aligned}
\mathbb E_Q(\widehat v-u)^2
&=\mathbb E_Q(\widehat v-v)^2+\mathbb E_Q(v-u)^2\\
&\quad+2\mathbb E_Q[(\widehat v-v)(v-u)].
\end{aligned}
\label{eq:targetdecomp}
\end{equation}
The cross term can have either sign; Appendix~\ref{app:extendedresults}
gives a concrete example.

\paragraph{Evaluation timing.}
Prediction uses the state in Table~\ref{tab:main-cohort}; structural
recovery always uses the retained pool after proposals. Both are
evaluated independently at each observation count, not as best-so-far
scores. Ground-truth forms are excluded from the discovery loop, and
held-out evaluation responses are excluded from all fitting and design.
The GT-containing diagnostic is specified separately in
Appendix~\ref{app:analysis-explicit}. Hypothetical acquisition responses
come from the fitted model, not additional simulator calls.

\paragraph{Curve aggregation.}
For method $a$, let $e_{a,j,s}(n)$ be the error of task $j$ and seed $s$,
and $S_j$ its included seeds. With $J$ tasks, define
\begin{equation}
\begin{aligned}
\ell_{a,j}(n)&=\frac{1}{|S_j|}\sum_{s\in S_j}\log e_{a,j,s}(n),\\
\bar\ell_a(n)&=\frac{1}{J}\sum_{j=1}^{J}\ell_{a,j}(n),\\
v_a(n)&=\frac{1}{J^2}\sum_{j=1}^{J}
\operatorname{Var}_{s\in S_j}[\log e_{a,j,s}(n)].
\end{aligned}
\label{eq:curve-aggregation}
\end{equation}
Lines show $\exp(\bar\ell_a)$; bands show
$\exp(\bar\ell_a\pm\sqrt{v_a})$. Variance is the sample variance
within each task (zero for a single included seed). These bands summarize
seed dispersion, not confidence intervals for paired improvements.
Each plotted point requires a prediction from every included run.
AutoSciLab's curves can therefore begin after the initial observation
budget because it needs additional observations before proposing its
first executable model. The fixed-pool comparison instead uses the
normalization and aggregation in Appendix~\ref{app:analysis-explicit}.

At the first ForceBench observation budget, predictions use the fitted
initial LLM proposals before any adaptive experiment is selected.
Performance at that point reflects the initial observations and fitted
pool, not an effect of an EIG-selected experiment.

\paragraph{Missing and early-terminated runs.}
BMD and ICL use all runs in Table~\ref{tab:main-cohort}. AutoSciLab's
predictive curves use 26/27 ChemBench and 8/15 BoxingGym runs; Box's
Apprentice uses 10/15 BoxingGym runs. All plotted methods use the full
ForceBench and NewtonBench sets. The exclusions are:
\begin{itemize}
\item AutoSciLab: ChemBench ping-pong medium, seed 37; BoxingGym death
(all three seeds), survival 61 (first law too late), hyperbolic 31
(overflow), and peregrines 31/41 (no finite values).
\item Box's Apprentice: death 41, dugongs 31, and all three hyperbolic
runs; hyperbolic 31 terminated at its cost cap.
\end{itemize}
Thus some curves summarize different task--seed sets. Five NewtonBench
AutoSciLab runs stopped with a confidence-based termination after
12--20 observations (Snell 31, Malus 31, sound speed 37, Hooke 31/37);
their last law is carried to $n=34$ for predictive evaluation.
Crashed BMD runs were rerun under the same seed with the cause recorded.
Missing predictions were not assigned zero error. Proposal rejection,
fit failure, and unavailable agent predictions are recorded separately.

The available-run counts above concern the sequential experiments only.
The 90/27/30/70 runs in Table~\ref{tab:explicit-summary} belong to the separate
fixed-pool diagnostic; they are not a revised denominator for
Figures~\ref{fig:main-results-new} and~\ref{fig:structural-recovery-main}.

\section{Numerical Implementation}
\label{app:method}\label{app:development-engine}

\paragraph{Model representation and validation.}
Each candidate specifies a functional expression and its coefficient
roles and domains. As in Section~\ref{sec:problem}, the observation-distribution
family is fixed for each task; it is not a hypothesis-search dimension.
The task interface supplies this likelihood and the link or forward model. Prediction uses the response-space conditional mean,
not the raw expression when the link is nonlinear.
ChemBench, BoxingGym, and NewtonBench allow
$+,-,\times,/,**$ and \texttt{exp}, \texttt{log}, \texttt{sqrt},
\texttt{tanh}, \texttt{square}, and \texttt{power}, with at most six
expression coefficients and fifteen operators. ForceBench additionally
allows \texttt{cos} and \texttt{sin}, but its engine enforces three
free constants and ten operators. Noise parameters are not counted as
expression coefficients.

Validation checks syntax, coefficient roles, consistency with the fixed
task likelihood, and finite predictions under 64 prior draws (on BoxingGym, at 16 probe
inputs). Canonicalization normalizes expression syntax, roles, and
family, including equivalent power notations. ForceBench also checks
numerical duplicates on a probe grid. These online duplicate tests are
not the full structural judge of Appendix~\ref{app:analysis-discovery}.
Rejected proposals and failed fits retain their failure reasons rather
than receiving an artificial evidence value.

\paragraph{Proposal and pool settings.}
The proposer uses Claude Opus 5.5 with medium reasoning effort,
16,000 output tokens, and a 120-second timeout. SDK retries are disabled;
a validation failure permits one repair turn. Initialization requests
12 candidates and permits two refill waves. Later rounds request three
candidates without refill. The pool retains up to 12 valid fitted model
families by evidence, with equal family-level prior weights. Every fifth round, the proposer may nominate up to two
retired candidates; each is refitted from its original prior on all
current data before competing for retention. All failed proposals and
retirements remain in the archive.

\begin{table*}[t]
\centering\small
\setlength{\tabcolsep}{4pt}
\caption{Benchmark-specific settings for the main BMD runs.
Common proposal and SMC settings are given in the text.
Appendices~\ref{app:analysis-feedback} and~\ref{app:analysis-joint}
specify analysis-specific differences.}
\label{tab:development-settings}
\begin{tabularx}{\textwidth}{@{}p{0.16\textwidth}*{4}{>{\raggedright\arraybackslash}X}@{}}
\toprule
& ChemBench & BoxingGym & ForceBench & NewtonBench\\
\midrule
Initial data & 3 fixed designs & 2 random designs & 2 trajectories & 4 random designs\\
Coefficient prior & Default & Default; hyperbolic: wide & Signed $\mathcal N(0,1)$; positive LogNormal$(0,1)$ & Wide\\
Observation family & Fixed LogNormal
& Fixed per task
& Fixed trajectory likelihood
& Fixed LogNormal\\
Archive review & Every 5 rounds to 30 & Every 5 rounds to 30 & Rounds 5, 10, 15 & Every 5 rounds to 30\\
EIG draws & 128 posterior draws & 128; count tasks use outcome tables & 128 outer and 128 inner draws & 128 posterior draws\\
Search starts & 5 uniform & 5 best candidate designs; survival enumerated & 5 uniform & 5 uniform\\
Optimizer & Projected finite differences; 20--500 steps
& Compass search; at most 40 steps
& Adam, learning rate 0.15; 40 steps
& As ChemBench\\
Paid-call cap & 160 & 100 & 100 & 160\\
\bottomrule
\end{tabularx}
\end{table*}

\paragraph{Role priors and observation models.}
ChemBench and BoxingGym's death, dugongs, peregrines, and survival tasks
use independent \emph{default} role priors:
\begin{equation}
\begin{aligned}
\theta_{\rm intercept}&\sim\mathcal N(0,2^2),\\
\theta_{\rm signed}&\sim\mathcal N(0,1),\\
\theta_{\rm positive}&\sim\operatorname{LogNormal}(0,1),\\
\theta_{\rm shape}&\sim\operatorname{Gamma}(2,1).
\end{aligned}
\label{eq:rolepriors}
\end{equation}
Gamma uses shape and rate. NewtonBench and BoxingGym's hyperbolic task
use a \emph{wide} prior: intercept $\mathcal N(0,4^2)$, signed
$\mathcal N(0,2^2)$, positive LogNormal$(0,2)$, and shape
Gamma$(2,\tfrac12)$. ForceBench uses the two priors in
Table~\ref{tab:development-settings}.

For a LogNormal response, let $\eta_m$ be the log-link expression and
$s_0$ the response unit computed from $\mathcal D_0$. Then
\begin{equation}
\begin{gathered}
\mu_m(\vx;\vtheta_m)=s_0\exp\{\eta_m(\vx;\vtheta_m)\},\\
Y\mid\vx,\vtheta_m,m\\[-2pt]
\sim\operatorname{LogNormal}(\log\mu_m-\sigma_m^2/2,\sigma_m),\\
\sigma_m\sim\operatorname{HalfNormal}(1).
\end{gathered}
\label{eq:chem_likelihood}
\end{equation}
Here $\mu_m$ is the response mean, not the median. The response unit is
the initial sample mean on ChemBench and BoxingGym's positive-response
path, and the initial geometric mean on NewtonBench. It remains fixed,
so evidence is conditional on this initial calibration.

ChemBench and NewtonBench use the fixed LogNormal observation family.
BoxingGym uses the likelihood assigned to each task; ForceBench uses
the task's fixed likelihood on trajectory coordinates. Candidates do
not change this distributional family, although parameters of the
specified likelihood can be inferred. The fitted likelihood can differ
from the simulator noise;
ChemBench, for example, generates relative Gaussian noise.
Equivalent expressions need not induce equivalent priors when the same
numerical role prior is used under a different parameterization.

\paragraph{SMC and prediction.}
SMC bridges densities proportional to
$\pi_m(\vtheta_m)L_{m,t}(\vtheta_m)^\beta$, increasing $\beta$ from
zero to one~\citep{delmoral2006smc}. Each refit starts from the original
prior and uses all current data. All benchmarks use 500 particles per
chain and two chains, with an effective-sample-size target of $0.5N$
and resampling at each tempering stage.
ChemBench, BoxingGym, and NewtonBench use PyMC 5.13.1, two worker cores,
and independent-Metropolis moves with adaptive move counts
(correlation threshold 0.01). Fit timeouts are 600 seconds on ChemBench
and NewtonBench; BoxingGym has no such timeout.

ForceBench uses JAX, runs the chains sequentially, and takes three
random-walk Metropolis moves per tempering stage. The fitting configuration
specifies a 5,000-stage/1,800-second limit.

For $C$ chains, the documented combination rule averages evidence
estimates, not their logarithms:
\begin{equation}
\log\widehat Z_{m,t}
=\operatorname{logsumexp}_{c=1}^{C}\log\widehat Z_{m,t}^{(c)}-\log C.
\label{eq:chaincombine}
\end{equation}
Posterior particles are concatenated with equal weights. With $N_m$
stored particles, the family predictive is
\[
q_{m,t}(y\mid\vx)\simeq\frac{1}{N_m}\sum_{k=1}^{N_m}
p_m(y\mid\vx,\vtheta_m^{(k)}).
\]
Point predictions average conditional response means over these
particles and then over families using evidence weights.

\paragraph{Joint-EIG estimation and search.}
The estimator samples a family, posterior parameters, and a hypothetical
outcome, then compares its conditional likelihood with the weighted
predictive mixture. Density calculations use log-sum-exp arithmetic.
Invalid components are tracked; an all-invalid query is a failure, not
zero information. Negative finite-sample estimates are not clipped.
The component estimators appear in Appendix~\ref{app:analysis-joint}.

ChemBench and NewtonBench start five searches uniformly in the scaled
unit box and select the best endpoint under the search draws. The
256-point pools declared for other comparison methods are not used by
these runs. ChemBench uses log coordinates for designated positive
inputs and affine coordinates otherwise. For iteration $k$, its
random-sign finite-difference schedules are
\[
a_k=\frac{0.2}{(10+k)^{0.602}},\qquad
c_k=\frac{0.05}{k^{0.101}},
\]
with clipped probes and updates, 20--500 steps, and a ten-step movement
window. BoxingGym scores a 100-point candidate pool (75 for hyperbolic)
and starts compass search from the best five points; it re-evaluates
the pool maximum and search endpoints together. Survival instead scores
every unobserved patient.

ForceBench uses five uniform starts and Adam ascent. One set of random draws is shared across
all starts and steps within a round; endpoints are re-estimated with
independent draws before selection. Every search respects the design
constraints and keeps the fitted model pool fixed.

\paragraph{EPIG implementation.}
The separate exploratory EPIG comparison in
Appendix~\ref{app:analysis-joint} uses the same predictive-mixture
construction and feasible domain. Its settings below describe that
auxiliary comparison, not a replacement for joint EIG in the main runs. BoxingGym samples 32 target inputs once per run from its
evaluation-input distribution, disjoint from evaluation points
(128 targets on hyperbolic; uniform resampling of the 100-patient pool
on survival). Search uses the same compass procedure, or enumeration
on survival. Continuous outcomes use 64 predictive-quantile intervals
including tails. Bounded counts are enumerated; unbounded counts are
enumerated to 100,000 with an aggregated tail at tolerance $10^{-12}$
and a 256 MiB working-memory limit. The resulting mutual information
is computed on finite probability tables.
ForceBench uses 32 uniform target designs, Adam search, and
$128\times128$ nested Monte Carlo draws without a quantile grid.
No evaluation responses are used in either acquisition.

\section{Strict Recovery and Ground-Truth Containment}
\label{app:analysis-discovery}

\paragraph{Criteria and selected models.}
The two curves in Figure~\ref{fig:structural-recovery-main} evaluate
the same top-1 model at each observation budget:
BMD's highest-evidence retained family or AutoSciLab's current hypothesis.
They neither search the archive for a better answer nor count earlier
successes cumulatively. BMD breaks evidence ties by the smallest
canonical model identifier. A missing AutoSciLab law counts as no
recovery; direct ICL has no structural output, so its recovery is not
applicable rather than zero.

For run $i$, let $\mathcal F_i^\star$ be the target family over its
admissible coefficient domain and $\mathcal F_i(m)$ the candidate family
on the benchmark's comparison scale. The intended relations are
\begin{equation}
\begin{aligned}
\text{strict: }&\mathcal F_i(m)=\mathcal F_i^\star,\\
\text{containment: }&\mathcal F_i^\star\subseteq\mathcal F_i(m).
\end{aligned}
\label{eq:appendix-structural-criteria}
\end{equation}
The plotted indicators record the outcomes of the checks below.
Containment includes strict success, so the curve gap counts accepted
containment cases without certified strict equivalence. Some of these
comparisons remain symbolically unresolved; the gap is not a count of
proven non-equivalent families.
Neither criterion requires recovery of the actual fitted coefficients or
of the observation-noise distribution.

String identity is unnecessary. For positive domains,
$ax/(b+x)=x/(\theta_1+\theta_2x)$ with
$\theta_1=b/a$ and $\theta_2=1/a$ is an invertible reparameterization.
By contrast, $c_2x^2+c_1x+c_0$ contains $ax+b$ by setting $c_2=0$,
but has a free quadratic term. The declared coefficient domains matter;
small estimated coefficients are not deleted after fitting.

\paragraph{Equivalence and containment checks.}
A deterministic judge (version 1.3) first applies the task's link, input
transformations, and target conventions. It uses neither an LLM nor the
agents' held-out prediction scores.
\emph{Strict equivalence} requires matching canonical expression trees,
a coefficient mapping solved in both directions, a 30-digit numerical
spot check, and admissible coefficients reproducing the true and
diagnostic parameter settings. Canonicalization splits exponentials of
sums and logarithms of products, merges powers, normalizes fractions,
extracts leading coefficients, and applies trigonometric addition
formulas. Diagnostic settings comprise the true setting plus two
additional settings on ChemBench, NewtonBench, and ForceBench, or three
on BoxingGym.

The judge marks \emph{non-equivalence} for incompatible input dependence,
coefficient-Jacobian rank, or coefficient domains, or for failure to
reproduce a target member within $10^{-3}$ under an eight-start refit.
\emph{Containment} is recorded separately. Its checks include up to
60 simplifying substitutions (coefficients set to $0$, $1$, or a
limiting $\infty$, or tied together), a solved mapping into a
higher-dimensional family, and wider coefficient domains. When no
symbolic witness is found, a numerical check fits diagnostic members
and requires error within $10^{-6}$ on separate check inputs.
This last check is evidence of numerical representability on those
inputs, not a symbolic proof of inclusion everywhere.

Symbolically unproved, numerically ambiguous, timed-out, or
non-identifiable cases remain \emph{unresolved}. They fail strict
recovery but can pass the separate containment check. Thus counting
strict certificates is conservative; failure to obtain a certificate does not itself
prove non-equivalence. These truth-assisted checks are evaluation-only and
never alter the discovered candidates or their fitted predictions.

\paragraph{Benchmark targets and exclusions.}
The counts below refer to the sequential recovery experiment in
Figure~\ref{fig:structural-recovery-main}. The separate fixed-pool study
uses the same notion of structural equivalence, with its own evaluation
cohort and denominators (Appendix~\ref{app:analysis-explicit}).

\emph{ChemBench} judges the primary kinetic law up to a positive overall
factor on the slice $\mathrm{pH}=7$ and $T=310\,\mathrm K$ (the latter
is not fixed for Arrhenius tasks). Candidate nuisance terms are recorded
but not judged. All 27 runs are included.

\emph{BoxingGym} judges the expression behind the link: $1-e^{-\theta x}$
for death, $a-bc^x$ for dugongs, the cubic log-mean for peregrines, and
$\lambda_0t e^{\beta m}$ on the logit scale for survival at both
treatment values. Hyperbolic requires the normal cumulative distribution
function, absent from the grammar, leaving 12 representable runs.

\emph{ForceBench} judges acceleration $F/p_2$ as a function of separation,
time, and the two parameters. Simulator force softening and periodic
background terms are excluded from the target; candidates adding them
can count as containment. Yukawa's Bessel function and the
extra-dimensions image sum are outside the interface, leaving 9 runs.

\emph{NewtonBench} judges the law after its log link, up to a positive
factor, with literal exponents fixed and named constants free. Malus
requires sine and cosine, absent from this grammar, leaving 24 runs.
Snell and Bose--Einstein use small-argument surrogate comparisons on
the tested input boxes: these can pass containment while remaining
unresolved for strict recovery. They are not exact-law certificates.
All excluded structural targets remain in predictive evaluation.

\paragraph{Agent schedules and interpretation.}
AutoSciLab uses every-round updates on ChemBench, BoxingGym, and
ForceBench, but its released NewtonBench schedule produces no law before
$n=16$. This differs from the every-round NewtonBench predictive curve.
Recovery uses the after-proposal model recorded at each budget. Averaging
per-seed-slot fractions equals pooling because each slot has the same
number of eligible runs. A correct earlier model can later be displaced,
so recovery can decrease. These curves alone do not distinguish failure
to propose a target family from failure to select it. Section~\ref{sec:c2}
addresses the latter question by including the GT family in every
candidate pool, without changing the shared-pool selection problem. The earlier primary-family classifier used in
Appendix~\ref{app:extendedresults} is broader than this strict criterion
and its results are reported separately.

\section{Numerical Evaluation on Shared Candidates}
\label{app:analysis-explicit}

\paragraph{Purpose and data.}
Section~\ref{sec:c2} separates explicit prediction, parameter inference,
and model selection from differences in hypothesis generation and
experimental design. For each world--seed pair, we collect 20
random-design observations, denoted by $\mathcal D$. This budget applies
to all four benchmarks in this diagnostic, including ForceBench; it is
separate from the sequential budgets in Table~\ref{tab:main-cohort}.
The reported cohort contains 90 ChemBench, 27 BoxingGym, 30 ForceBench,
and 70 NewtonBench runs, for a total of 217. A--D share the same data
and evaluation inputs within each run. No additional observations or
adaptive experiments are acquired during this comparison.

\paragraph{Candidate construction.}
Given only the first 10 observations, Claude Opus 5.5 proposes 11
alternative model families. We combine them with the ground-truth (GT)
family to form a fixed candidate list $\mathcal L$ of size $K=12$.
The GT insertion makes the correct functional form available for
selection; it does not select that family on behalf of any method.
All methods then receive the same list and use all 20 observations.
The pool can differ between world--seed pairs but is fixed across A--D
within a pair. Likelihood families are fixed per task, as in
Section~\ref{sec:problem}.

\paragraph{LLM prediction and selection.}
A predicts responses directly from the observations and candidate list.
B jointly selects a family $m_L\in\mathcal L$ and estimates its
parameters $\widetilde{\vtheta}_{m_L}$; the resulting model is executed
at the evaluation inputs. A and B do not use SMC parameter estimates,
model evidence, or residual feedback. A receives prediction queries,
whereas B produces an executable model. Thus A--B compares direct
numerical prediction with prediction through an explicit model, not
an isolated change to the LLM's internal representation.

\paragraph{Shared fits and predictions.}
The numerical engine fits every candidate on the full $\mathcal D$
using SMC and the task priors, independently of B's parameter estimates.
For a candidate $m$, define its conditional response mean
$\mu_m(\vx;\vtheta_m)=\mathbb E[Y\mid\vx,\vtheta_m,m]$ and its
posterior predictive mean
\begin{equation}
\overline\mu_m(\vx)=
\mathbb E_{\vtheta_m\mid\mathcal D,m}[\mu_m(\vx;\vtheta_m)].
\label{eq:c2-posterior-mean-prediction}
\end{equation}
Let $\widehat Z(m)$ denote its estimated marginal likelihood and let
$m_E\in\arg\max_{m\in\mathcal L}\log\widehat Z(m)$ be the
evidence-selected family. The three explicit predictors are
\begin{equation}
\begin{aligned}
\widehat\mu_{\rm B}(\vx)&=\mu_{m_L}(\vx;\widetilde{\vtheta}_{m_L}),\\
\widehat\mu_{\rm C}(\vx)&=\overline\mu_{m_L}(\vx),\\
\widehat\mu_{\rm D}(\vx)&=\overline\mu_{m_E}(\vx).
\end{aligned}
\label{eq:c2-fixed-predictions}
\end{equation}
C retains B's selected family. C and D reuse the same fitted candidates
and posterior predictions, so matching family selections give identical
predictions. Posterior prediction averages conditional response means,
not parameters before nonlinear execution. B--C therefore compares
within-family inference, while C--D changes only family selection.
D uses one family; it is neither Bayesian model averaging nor a run of
the complete BMD discovery loop.

\paragraph{Prediction targets.}
All four methods use the same held-out inputs, targets, and scorer
within each run. ChemBench and NewtonBench use 256 evaluation inputs
with nMSLE; BoxingGym uses 256 inputs with nMSE. ForceBench uses nMSE
over two-dimensional trajectories at 10 measurement times under six
evaluation designs. The response targets and metric normalizers follow
Appendix~\ref{app:analysis-common}. In particular, the BoxingGym summary
uses nMSE throughout, rather than mixing nMSE and Brier scores.

\paragraph{Error aggregation and normalization.}
For benchmark $b$, let $W_b$ be its number of evaluated worlds,
$S_{b,j}$ the included seeds of world $j$, and $e_{a,b,j,s}$ the
prediction error of method $a$. The benchmark aggregate is
\begin{equation}
\begin{aligned}
\ell_{a,b,j}
&=\frac{1}{|S_{b,j}|}\sum_{s\in S_{b,j}}
  \log\max(e_{a,b,j,s},10^{-6}),\\
E_{a,b}
&=\exp\!\left\{\frac{1}{W_b}\sum_{j=1}^{W_b}\ell_{a,b,j}\right\}.
\end{aligned}
\label{eq:c2-geometric-error}
\end{equation}
Errors are first geometrically averaged over seeds within each world,
then over worlds. Table~\ref{tab:explicit-summary} reports relative errors
$R_{a,b}=E_{a,b}/E_{{\rm A},b}$, making direct ICL equal to one in
every benchmark. The overall relative error is
\begin{equation}
R_a^{\rm overall}=
\exp\!\left\{\frac14\sum_{b=1}^{4}\log R_{a,b}\right\}.
\label{eq:c2-overall-relative}
\end{equation}
This gives equal weight to each benchmark, not to every run across all
benchmarks. Table~\ref{tab:c2-absolute} gives the reported absolute
errors before normalization, and Table~\ref{tab:c2-fixed-audit} gives
the identification counts. Values are shown at the precision supplied
in the result summaries.

\begin{table*}[t]
\centering\small
\setlength{\tabcolsep}{8pt}
\renewcommand{\arraystretch}{1.1}
\caption{Absolute prediction errors for the shared-candidate study in
Table~\ref{tab:explicit-summary}, before normalization by A. Entries are
geometric means over seeds within worlds and then over worlds;
the overall row is the reported geometric mean across benchmarks.
Lower is better.}
\label{tab:c2-absolute}
\begin{tabularx}{\textwidth}{@{}Xcccc@{}}
\toprule
Benchmark (metric) & A & B & C & D\\
\midrule
ChemBench (nMSLE) & $8.47\!\times\!10^{-4}$ & $7.77\!\times\!10^{-4}$ & $2.97\!\times\!10^{-4}$ & $2.96\!\times\!10^{-4}$\\
BoxingGym (nMSE) & $4.17\!\times\!10^{-2}$ & $3.62\!\times\!10^{-2}$ & $5.79\!\times\!10^{-2}$ & $2.78\!\times\!10^{-2}$\\
ForceBench (nMSE) & $1.00\!\times\!10^{-3}$ & $2.30\!\times\!10^{-4}$ & $1.74\!\times\!10^{-4}$ & $3.58\!\times\!10^{-5}$\\
NewtonBench (nMSLE) & $9.34\!\times\!10^{-6}$ & $7.26\!\times\!10^{-6}$ & $2.05\!\times\!10^{-6}$ & $2.31\!\times\!10^{-6}$\\
\midrule
Overall (geomean) & $7.58\!\times\!10^{-4}$ & $4.65\!\times\!10^{-4}$ & $2.80\!\times\!10^{-4}$ & $1.62\!\times\!10^{-4}$\\
\bottomrule
\end{tabularx}
\end{table*}

\begin{table}[t]
\centering\small
\setlength{\tabcolsep}{7pt}
\renewcommand{\arraystretch}{1.1}
\caption{Top-1 GT identification in the fixed-pool study
(correct / evaluated runs). B and C select the same family.
A has no structural output.}
\label{tab:c2-fixed-audit}
\begin{tabular}{@{}lcc@{}}
\toprule
Benchmark & B/C & D\\
\midrule
ChemBench & 89/90 & 87/90\\
BoxingGym & 5/27 & 17/27\\
ForceBench & 15/30 & 25/30\\
NewtonBench & 63/70 & 55/70\\
\midrule
Overall & 172/217 & 184/217\\
\bottomrule
\end{tabular}
\end{table}

\paragraph{Top-1 GT identification.}
A hit requires structural equivalence of the selected family to the GT,
using the criterion in Appendix~\ref{app:analysis-discovery}; merely
containing the GT as a special case is not sufficient. B and C have
identical identification outcomes because they select the same family.
D is evaluated using its highest-evidence family. Identification is
not applicable to A, which does not return a selected model.
The overall rate pools all 217 runs: B/C identify the GT in 172/217
runs (79.3\%), and D in 184/217 (84.8\%). Unlike the error aggregate,
this overall rate is weighted by each benchmark's number of runs.

\paragraph{Interpretation.}
The cross-benchmark error decreases from A to B to C to D. The paired
comparisons support explicit model execution over direct ICL, Bayesian
fitting within the selected family, and evidence-based comparison of
fitted families, respectively. Together, they support the division of
labor described in Section~\ref{sec:c2}. The GT-containing pool makes
this a test of model evaluation when a correct candidate is available,
rather than a test of generating that candidate or choosing experiments.

\section{Belief and Diagnostic Feedback}
\label{app:analysis-feedback}

\paragraph{Purpose and conditions.}
Section~\ref{sec:c3} tests whether numerical feedback helps the LLM
propose subsequent hypotheses, beyond the numerical engine's role in
fitting and comparing existing models. Let $b,d\in\{0,1\}$ indicate
whether belief and diagnostic feedback are exposed to the proposer.
The $2\times2$ ablation compares Full $(1,1)$, Belief-only $(1,0)$,
Diagnostics-only $(0,1)$, and No feedback $(0,0)$.
All variants retain parameter inference, evidence weighting, Bayesian
model averaging, candidate generation, and joint-EIG design. Only the
feedback available to the proposer changes; ``No feedback'' does not
remove the numerical engine.

\begin{table*}[t]
\centering\small
\caption{Information shown to the proposer in the feedback ablation.
Numerical inference remains active in all four variants.}
\label{tab:feedback-fields}
\begin{tabularx}{\textwidth}{@{}p{0.16\textwidth}Xp{0.14\textwidth}@{}}
\toprule
Bundle & Contents & Visible in\\
\midrule
Common & Observations, task interface, candidate expressions, coefficient
roles, grammar, structural duplicate history, syntax/repair status,
and eligible retired identifiers & All variants\\
Belief & Log evidence, model weights, best-model marker, parameter
posterior summaries, pre-observation predictive log scores, and archived
evidence values & $b=1$\\
Diagnostics & Standardized predictive residuals for every active model
at the acquired observations, identified by model & $d=1$\\
\bottomrule
\end{tabularx}
\end{table*}

\paragraph{Feedback masking.}
Table~\ref{tab:feedback-fields} specifies the feedback bundles.
Candidates and archive entries are listed in canonical-identifier order.
Removing belief also removes evidence rankings, best-model annotations,
and evidence-based retirement explanations from structured fields,
prose, and conversation history. Retired-model evidence values remain
only when belief is visible. The composition and archive-review
instructions use neutral wording (Appendix~\ref{app:prompts}).
Diagnostics provide model-wise residuals rather than selecting only a
preferred model. Removing a bundle does not remove the observed data,
candidate identities, or active-pool membership.

One-step predictive scores are computed from predictions made before
the corresponding response is observed. Residuals use each model's
predictive scale. Even with explicit belief fields removed, residuals
can reveal relative fit quality, and pool membership reflects numerical
selection. The ablation therefore tests these feedback bundles, not the
elimination of all indirect evidence about candidate quality.

\paragraph{Initialization and execution.}
Figure~\ref{fig:joint-analysis} compares nine ChemBench tasks and five
ForceBench worlds, each with three seeds (27 and 15 runs per variant).
Within each task--seed group, all four variants start from the same
initial observations $\mathcal D_0$, initial candidate pool, and fitted
state, including parameter posteriors and evidence estimates. A saved
round-0 state is replayed to keep this initialization identical.
From the first feedback-guided proposal onward, each variant follows
its own trajectory: it proposes families, chooses experiments, obtains
observations, and updates its fitted pool. Likelihoods, priors, grammar,
pool capacity, proposal counts, and archive-review eligibility are
otherwise kept fixed.

This is a sequential feedback ablation, not the fixed-data comparison
of Section~\ref{sec:c2}. Subsequent datasets and model pools need not
match, even though initialization and the numerical inference and
design procedures are shared.

\paragraph{Evaluation and interpretation.}
The horizontal axis is the total number of observations. ChemBench
uses observation-mean nMSLE, and ForceBench uses trajectory nMSE on
the shared evaluation designs. The target definitions and geometric
aggregation follow Appendix~\ref{app:analysis-common}. The ChemBench
panel displays observations 13--33; ForceBench displays 2--18.

In Figure~\ref{fig:joint-analysis}, Full has the lowest final error on
both benchmarks, while No feedback has the highest. Either feedback
bundle generally improves prediction over No feedback. These results
support returning numerical beliefs and residual diagnostics to the
LLM, rather than using numerical inference only to score its existing
hypotheses. Because proposals and observations subsequently evolve,
the measured effect is on downstream prediction in the discovery loop,
not an isolated score of individual proposal quality. The plotted bands
summarize seed dispersion, not confidence intervals for paired gains.

\section{Joint EIG and Its Information Components}
\label{app:analysis-joint}

\paragraph{Objectives and estimator.}
With a fixed dataset and fitted pool, the compared objectives are
\begin{equation}
\begin{aligned}
a_{{\rm J},t}(\vx)&=I(Y_\vx;M,\boldsymbol\Theta_M\mid\mathcal D_t),\\
a_{{\rm M},t}(\vx)&=I(Y_\vx;M\mid\mathcal D_t),\\
a_{{\rm P},t}(\vx)&=\sum_m w_{m,t}
 I(Y_\vx;\boldsymbol\Theta_m\mid\mathcal D_t,M=m).
\end{aligned}
\label{eq:analysis-design-terms}
\end{equation}
Parameter-only averages over all families; Model-only uses
parameter-integrated predictions, not point estimates. Random samples
from the declared design distribution.
For $S$ shared draws $(m_s,\vtheta_s,y_s)$, define
$r_s=p_{m_s}(y_s\mid\vx,\vtheta_s)$,
$u_s=q_{m_s,t}(y_s\mid\vx)$, and $v_s=q_t(y_s\mid\vx)$.
The estimates are
\begin{equation}
\begin{aligned}
\widehat a_{{\rm J},t}&=S^{-1}\sum_s\log(r_s/v_s),\\
\widehat a_{{\rm M},t}&=S^{-1}\sum_s\log(u_s/v_s),\\
\widehat a_{{\rm P},t}&=S^{-1}\sum_s\log(r_s/u_s).
\end{aligned}
\label{eq:appendix-shared-eig}
\end{equation}
Identical draws and density evaluations preserve
$\widehat a_{\rm J}=\widehat a_{\rm M}+\widehat a_{\rm P}$ up to
arithmetic error. Separately clipping estimates or removing different
invalid components breaks this identity; the identity itself does not
establish convergence.

\paragraph{Prediction-oriented EPIG.}
Expected predictive information gain (EPIG) targets responses at inputs
of interest
~\citep{bickfordsmith2023epig}. Let $\vx'\sim\nu$ be a feasible target
input sampled as specified in Appendix~\ref{app:method}. The prospective
response at $\vx$ and target response at $\vx'$ share the same sampled
family and parameters but are conditionally independent given them:
\begin{equation}
\begin{aligned}
&q_t^{(2)}(y,y'\mid\vx,\vx')\\
&=\mathbb E_{M,\boldsymbol\Theta_M}\!\left[
 p_M(y\mid\vx,\boldsymbol\Theta_M)
 p_M(y'\mid\vx',\boldsymbol\Theta_M)\right].
\end{aligned}
\label{eq:bmd-joint-response-predictive}
\end{equation}
Then
\begin{equation}
\begin{aligned}
a_{\mathrm{EPIG},t}(\vx)
&=\mathbb E_{\vx'\sim\nu} I(Y_\vx;Y'_{\vx'}\mid\mathcal D_t)\\
&=\mathbb E\log\frac{q_t^{(2)}(y,y'\mid\vx,\vx')}
{q_t(y\mid\vx)q_t(y'\mid\vx')}.
\end{aligned}
\label{eq:bmd-epig}
\end{equation}
The second expectation samples $\vx'\sim\nu$ and then
$(y,y')\sim q_t^{(2)}$. Independent latent draws would remove the
predictive dependence of interest. These responses are hypothetical;
EPIG does not query held-out labels.

\paragraph{Current joint-EIG comparison.}
Figure~\ref{fig:feedback} compares Joint EIG, Model-only EIG,
Parameter-only EIG, and Random on ChemBench and ForceBench, as described
in Section~\ref{sec:c4}. The comparison uses nine ChemBench tasks and
five ForceBench worlds with three seeds each. Candidate generation,
Bayesian fitting, evidence weighting, and posterior prediction are
retained in every variant; the experimental-design objective changes.
The parameter-only objective averages within-family information using
the same model weights, rather than conditioning on a single selected
family.

Each acquisition policy chooses designs in the feasible space and
updates its own observations and candidate pool. Holding the fitted
state fixed while optimizing one next experiment should not be confused
with fixing that state for the entire sequential run. The ChemBench
panel displays observations 13--33, and ForceBench displays 2--18;
targets and geometric aggregation follow
Appendix~\ref{app:analysis-common}.
The figure supports the benefit of jointly targeting model and parameter
uncertainty: Joint EIG maintains lower error through much of the later
ChemBench budgets and reduces ForceBench error faster at intermediate
budgets, with closer performance at the end.

\paragraph{Separate exploratory EPIG comparison.}
The supplied auxiliary acquisition results also compare EPIG with the
EIG objectives. Table~\ref{tab:acquisition-finals} reports these earlier
runs separately: 27 ChemBench, 15 BoxingGym, and 15 ForceBench runs
per evaluated objective. EPIG was evaluated on BoxingGym and ForceBench,
not ChemBench. This auxiliary study uses the earlier feedback
configuration, with residuals for the best model, evidence-ordered
candidates, and the original composition and archive instructions.
Its ForceBench runs predate the later SMC fit limit. These are not the
runs plotted in Figure~\ref{fig:feedback} or
Figure~\ref{fig:joint-analysis}.

Within this auxiliary study, likelihood settings, priors, proposal
rules, fitting resources, and design-search settings are held fixed
apart from the acquisition objective. Each variant generates its own
initial pool, so the realized pools, subsequent data, and proposals
need not coincide. The values below are retained as the supplied
separate EPIG evaluation, not as final values for the current
joint-EIG figure or as measurements from the unified baseline rerun.

\begin{table}[t]
\centering\small
\setlength{\tabcolsep}{4pt}
\caption{Earlier auxiliary acquisition comparison, separate from
Figure~\ref{fig:feedback}. Entries are final geometric-mean errors;
ChemBench uses $n=33$, BoxingGym $n=32$, and ForceBench $n=18$.
The feedback and initialization configuration is described in the text.}
\label{tab:acquisition-finals}
\begin{tabular}{@{}lccc@{}}
\toprule
Objective & ChemBench & BoxingGym & ForceBench\\
\midrule
Joint EIG & $1.47\!\times\!10^{-5}$ & 0.0101 & $8.8\!\times\!10^{-8}$\\
Model-only & $1.93\!\times\!10^{-5}$ & 0.0111 & $8.2\!\times\!10^{-6}$\\
Parameter-only & $1.63\!\times\!10^{-5}$ & 0.0168 & $3.6\!\times\!10^{-6}$\\
Random & $2.98\!\times\!10^{-5}$ & 0.0570 & $3.6\!\times\!10^{-5}$\\
EPIG & --- & 0.0132 & $9.5\!\times\!10^{-7}$\\
\bottomrule
\end{tabular}
\end{table}

In this auxiliary comparison, EPIG has lower final error than Random
on BoxingGym and ForceBench, while Joint EIG has lower error than EPIG.
These results supplement the comparison of information objectives;
they should not be used to reconstruct the curves or effect sizes of
the current main experiments. No NewtonBench acquisition result is
reported.

\section{Proposer Prompt Components}
\label{app:prompts}\label{app:u3-protocol}

The proposer combines the supplied full-feedback base instruction
(FA--C-R) with the three-candidate and archive-review extensions below.
Each request also supplies the task interface, observations, feedback,
and task-specific constraints. The base asks for one candidate; the
composition extension overrides this output format with
\texttt{\{limit\}} set to three. FA--C-R provides residuals for all
active models. The auxiliary comparison in
Appendix~\ref{app:analysis-joint} uses the older FA--C base and earlier
extensions, which differ from the current full-feedback configuration.

\paragraph{Task constraints and terminology.}
In the supplied prompt schema, the field named \texttt{family} denotes
the observation-distribution family, whereas ``model family'' in the
paper denotes the functional form. The experiments described in
Section~\ref{sec:problem} hold the observation-distribution family
fixed per task. Accordingly, the schema's conditional support for
\texttt{family\_expansion} does not make likelihood-family search part
of the reported method: this flag is disabled, and \texttt{family}
must match the task's fixed distribution.

Task constraints also specify coefficient roles, expression operators,
and the link or forward model. ForceBench includes \texttt{cos} and
\texttt{sin}; the engine applies its task-specific limits of three
free constants and ten operators. The generic base text below includes
the six-coefficient/fifteen-operator limits used for the other interfaces.
The task-specific constraint block and validation enforce the applicable
limits. Parameter priors are assigned by coefficient role rather than
written numerically by the LLM.

\paragraph{Initialization and feedback variants.}
Discovery initialization requests candidate forms from the initial
observations, with the candidate/refill settings in
Appendix~\ref{app:method}. The GT-containing pool of
Section~\ref{sec:c2} is a separate controlled construction, not a prompt
used to initialize the end-to-end discovery runs. In the feedback
ablation, all four variants replay the same initial fitted state before
receiving their respective feedback bundles.
Belief-only removes model-wise residuals; Diagnostics-only removes
belief fields and their ranking cues; No feedback removes both bundles.
The masking covers structured fields, accompanying prose, and history
as specified in Appendix~\ref{app:analysis-feedback}. The observed data,
candidate identities, and modeling interface remain available.

\par\noindent\textbf{Full-feedback base prompt.}\par\smallskip
\begin{nolinenumbers}
\begin{lstlisting}[basicstyle=\ttfamily\small,
 breaklines=true,breakatwhitespace=false,columns=fullflexible,
 frame=single,aboveskip=5pt,belowskip=5pt]
# prompt_version: v2.9.FA-C-R
Propose one new candidate model for the process described only by the data and the structured fields in the user message. A Bayesian engine fits and scores every candidate.

The user message is a JSON belief packet: interface (outcome type and support, input names and ranges, the fixed link), observations (all public data so far), pool (each active model with expression, family, evidence weight, log evidence, one-step predictive log-score history, and posterior mean and sd of each coefficient), residuals_by_model (for every active model, the standardized residual of each public observation under that model's posterior predictive, keyed by model id and sorted by input), archive (every earlier proposal in this run with its outcome), and constraints. Nothing in the packet comes from held-out data. A large fitted dispersion makes standardized residuals look small even when the mean form is wrong.

Propose the one (expression, family) pair that you judge most likely to have generated all of the observations and that is not in the pool or the archive; if your first choice is already there, propose your next. Judge a form by how much of the data it would leave unexplained, not by how few edits separate it from the current best model. Inputs may combine through products, ratios or nested functions as well as through sums of separate terms.

Link
- Under a log link, expression is log(mean / response_unit): a factor f of the mean enters expression as + log(f) and a divisor g as - log(g). Inside log a raw input with a wide positive range is safe; outside log, such an input times a coefficient, or raised to a coefficient power, can overflow exp and is rejected.

Checks
- predicted_diagnostic: where your model should fit the observations better than the current best model, as {"input": <one input name>, "low": <number>, "high": <number>}. The two fitted models are compared on the observations whose value of that input lies between low and high.
- closest_archived_model: the expression in the pool or the archive that is most similar to yours, copied exactly.

Rules
- expression is the argument of the fixed link. Use only coefficient names declared in parameters, the input names given in interface, real literals, the operators + - * / **, and exp, log, sqrt, tanh, square, power(x, k). At most 6 coefficients and 15 operators.
- Count operators as the engine does: each +, -, *, /, ** and each unary minus counts 1, and each call of exp, log, sqrt, tanh, square or power counts 1; names and numbers count 0. For example c*tanh(w) counts 2, and each further term c*f(x + k) added to a sum costs 4. If your count is above 15, simplify the expression before replying.
- parameters maps every coefficient used in expression to exactly one of: intercept, signed, positive, shape (at most one intercept). Priors are fixed by these labels; never write numeric priors. Family dispersion parameters belong to the family; never list them.
- family must be one of constraints.allowed_families. If constraints.structure_expansion is false, expression must be verbatim one of constraints.initial_structures and only family may change. If constraints.family_expansion is false, family must equal constraints.initial_family.
- Your whole reply is one raw JSON object - its first character is { and its last is } - with the keys {"expression": ..., "family": ..., "parameters": {...}, "predicted_diagnostic": {...}, "closest_archived_model": ...}: no code fence and no text before or after it. If it fails validation you receive the error once and may reply once more.
\end{lstlisting}
\end{nolinenumbers}

\par\noindent\textbf{Three-candidate extension.}\par\smallskip
\begin{nolinenumbers}
\begin{lstlisting}[basicstyle=\ttfamily\small,
 breaklines=true,breakatwhitespace=false,columns=fullflexible,
 frame=single,aboveskip=5pt,belowskip=5pt]
Return a JSON array of exactly {limit} candidate objects, each with exactly the keys specified above and nothing else. Propose three distinct, non-equivalent hypotheses. At least two of the three candidates should differ meaningfully in their primary functional composition, rather than merely in parameterization or by adding or removing a correction term. At most one candidate may be a nested or incremental refinement of a model already in the pool, and only if the public observations specifically call for that refinement.
\end{lstlisting}
\end{nolinenumbers}

\paragraph{Archive-review extension.}
Every fifth round, the extension below also allows up to two retired
model IDs. Each nominated model is refitted from its prior on current
data, rather than restored with an old evidence value.
\begin{nolinenumbers}
\begin{lstlisting}[basicstyle=\ttfamily\small,
 breaklines=true,breakatwhitespace=false,columns=fullflexible,
 frame=single,aboveskip=5pt,belowskip=5pt]
This is an archive review round. Besides your {limit} new candidates, you may ask for up to two archived models to be re-fit: choose only among archive entries whose outcome is retired and whose form you now judge likely given all current public observations. Each model you name is re-fit from its prior on all current public data and takes part in the pool like any other model. Reply with one raw JSON object {"candidates": [...], "revive": [...]}: candidates is the array described above, and revive lists the chosen archive model_id strings exactly as given, or is empty when nothing in the archive deserves another fit.
\end{lstlisting}
\end{nolinenumbers}

\section{Earlier ChemBench Diagnostics}
\label{app:extendedresults}\label{app:development-reconstruction}
\label{app:poolhistories}

\paragraph{Scope.}
These earlier ChemBench diagnostics examine prediction targets,
parameter fitting, and calibration in a separate development cohort.
They are not the runs underlying Figures~\ref{fig:main-results-new}--\ref{fig:joint-analysis}
or Table~\ref{tab:explicit-summary}, and their longer budgets, aggregation,
and supplied-form controls should not be substituted for the current
experimental protocols.
They cover Michaelis--Menten (c0), product inhibition (c2), and ping-pong
bi-substrate (c5), each at three difficulties, with seeds 31, 37, and 41
and three ICL repetitions. Runs use three initial and 60 additional observations ($n=63$).
Task tables use 256 evaluation inputs; detailed ping-pong reconstructions
and supplied-form controls use the official 1,000-input set.

\paragraph{Prediction target and aggregation.}
Table~\ref{tab:chembench-final-tasks} gives arithmetic seed means for
each task. Development aggregates take their geometric mean, unlike
the main figures, which average log errors over seeds first.
\begin{table*}[tp]
\centering
\small
\setlength{\tabcolsep}{5pt}
\caption{Earlier ChemBench diagnostic nMSLE at $n=63$ on 256 evaluation inputs. These are not per-task values for the current main figure.}
\label{tab:chembench-final-tasks}
\begin{tabular}{lrrrrrr}
\toprule
Task & Primary EIG & Primary random & Primary ICL & Obs. EIG & Obs. random & Obs. ICL \\
\midrule
c0 easy & $2.13\!\times\!10^{-4}$ & $2.14\!\times\!10^{-4}$ & $2.08\!\times\!10^{-4}$ & $1.69\!\times\!10^{-6}$ & $1.11\!\times\!10^{-5}$ & $1.96\!\times\!10^{-4}$ \\
c0 medium & $2.18\!\times\!10^{-4}$ & $2.79\!\times\!10^{-4}$ & $2.29\!\times\!10^{-5}$ & $2.88\!\times\!10^{-6}$ & $1.36\!\times\!10^{-5}$ & $2.45\!\times\!10^{-4}$ \\
c0 hard & $2.17\!\times\!10^{-4}$ & $2.46\!\times\!10^{-4}$ & $2.99\!\times\!10^{-5}$ & $5.99\!\times\!10^{-6}$ & $8.62\!\times\!10^{-6}$ & $2.81\!\times\!10^{-4}$ \\
c2 easy & $1.95\!\times\!10^{-4}$ & $2.01\!\times\!10^{-4}$ & $3.65\!\times\!10^{-4}$ & $5.78\!\times\!10^{-6}$ & $6.26\!\times\!10^{-6}$ & $6.62\!\times\!10^{-5}$ \\
c2 medium & $2.01\!\times\!10^{-4}$ & $2.25\!\times\!10^{-4}$ & $1.72\!\times\!10^{-4}$ & $8.72\!\times\!10^{-7}$ & $1.22\!\times\!10^{-5}$ & $2.02\!\times\!10^{-4}$ \\
c2 hard & $1.94\!\times\!10^{-4}$ & $1.85\!\times\!10^{-2}$ & $2.29\!\times\!10^{-4}$ & $6.71\!\times\!10^{-6}$ & $1.82\!\times\!10^{-2}$ & $1.57\!\times\!10^{-4}$ \\
c5 easy & $2.41\!\times\!10^{-4}$ & $2.18\!\times\!10^{-4}$ & $3.69\!\times\!10^{-3}$ & $9.35\!\times\!10^{-6}$ & $1.26\!\times\!10^{-5}$ & $3.56\!\times\!10^{-3}$ \\
c5 medium & $2.16\!\times\!10^{-4}$ & $2.30\!\times\!10^{-4}$ & $3.60\!\times\!10^{-4}$ & $9.88\!\times\!10^{-6}$ & $1.18\!\times\!10^{-5}$ & $3.18\!\times\!10^{-4}$ \\
c5 hard & $2.41\!\times\!10^{-4}$ & $2.43\!\times\!10^{-4}$ & $6.68\!\times\!10^{-3}$ & $8.67\!\times\!10^{-6}$ & $2.72\!\times\!10^{-5}$ & $6.78\!\times\!10^{-3}$ \\
\bottomrule
\end{tabular}
\par\smallskip
\begin{minipage}{0.97\linewidth}\footnotesize Entries are arithmetic means of three algorithm seeds or ICL replicates. c0 is Michaelis--Menten, c2 product inhibition, and c5 ping-pong bi-substrate. Algorithm seeds do not change the ChemBench truth world in this cohort.\end{minipage}
\end{table*}

Observation-target aggregates are EIG $4.52\times10^{-6}$, Random
$2.69\times10^{-5}$, and ICL $3.92\times10^{-4}$, with EIG lowest on
all nine tasks. Primary-target aggregates are $2.15\times10^{-4}$,
$3.76\times10^{-4}$, and $2.97\times10^{-4}$, with the lowest error
on 4, 1, and 4 tasks, respectively.
EIG improves observation error over Random in 19/27 seed comparisons,
not every seed. Removing the product-inhibition hard task in a recorded
sensitivity check changes the EIG/Random ratio from 0.571 to 0.941 for
the primary target and from 0.168 to 0.362 for the observation target.
The task remains included.

\paragraph{A discovered family and its fitted predictor.}
Table~\ref{tab:appendix-c5-traces} records first admission of a
ping-pong primary family. This older classifier accepts additional
nuisance terms and is a containment diagnostic, not the current strict
judge. For EIG seed 31, the round-9 expression ($n=12$) is
\begin{equation}
\begin{aligned}
\eta(\vx)&=a+\log X_5+\log X_1+\log X_3\\
&\quad-\log\{KX_3+X_1(X_3+K_a)\},
\end{aligned}
\label{eq:entryexample}
\end{equation}
where $X_1,X_3$ are substrate inputs and $X_5$ is enzyme concentration;
its link is Equation~\eqref{eq:chem_likelihood}. Posterior means
$K=1.032$ and $K_a=2.060$ are close to true values 1 and 2, with
particle standard deviations 0.061 and 0.140. Family weight exceeds
0.9999996 at first admission in all six histories.
\begin{table*}[t]
\centering\small
\setlength{\tabcolsep}{4pt}
\caption{Earlier ping-pong-easy fitted histories on the official 1,000-input set.
Entry and stable dominance concern the broader primary-family classifier.
All errors are nMSLE. Final values use $n=63$; admission rounds count
observations acquired after $n_0=3$.}
\label{tab:appendix-c5-traces}
\begin{tabular*}{\textwidth}{@{\extracolsep{\fill}}llrrrrrr@{}}
\toprule
& & Entry & Stable & \multicolumn{2}{c}{Primary error at admission}
& \multicolumn{2}{c}{Final error}\\
\cmidrule(lr){5-6}\cmidrule(l){7-8}
Policy & Seed & round & round & Before & After & Primary & Observation\\
\midrule
EIG & 31 & 9 & 9 & $8.18\!\times\!10^{-1}$ & $1.64\!\times\!10^{-4}$ & $2.72\!\times\!10^{-4}$ & $2.90\!\times\!10^{-6}$\\
EIG & 37 & 7 & 7 & $8.48\!\times\!10^{-1}$ & $4.84\!\times\!10^{-4}$ & $2.86\!\times\!10^{-4}$ & $1.29\!\times\!10^{-5}$\\
EIG & 41 & 18 & 18 & $3.12\!\times\!10^{-3}$ & $4.65\!\times\!10^{-5}$ & $2.78\!\times\!10^{-4}$ & $1.04\!\times\!10^{-5}$\\
Random & 31 & 9 & 9 & $1.31\!\times\!10^{-1}$ & $9.29\!\times\!10^{-5}$ & $2.76\!\times\!10^{-4}$ & $1.05\!\times\!10^{-5}$\\
Random & 37 & 5 & 17 & $5.65\!\times\!10^{-1}$ & $5.27\!\times\!10^{-5}$ & $2.29\!\times\!10^{-4}$ & $1.39\!\times\!10^{-5}$\\
Random & 41 & 7 & 7 & $2.66\!\times\!10^{-1}$ & $2.07\!\times\!10^{-4}$ & $2.55\!\times\!10^{-4}$ & $1.19\!\times\!10^{-5}$\\
\bottomrule
\end{tabular*}
\end{table*}

A reconstruction of 26 saved fits reproduces primary scores within
$1.11\times10^{-16}$. Admission coincides with new data and refitting,
so error changes do not isolate its causal effect. The illustration
uses the lowest seed, not a random success. Stable dominance requires
total classified-family weight to remain at least 0.5 thereafter, not
continued retention of one expression: the first EIG seed-31 model
retires at round 28. Across six histories, 46--66 models enter the pool
and the top-weight model changes 29--44 times. All 54 runs in the larger
development set end with 12 active models and none of their original
models. These counts describe pool turnover, not additional recoveries.

On the 1,000-input ping-pong-easy set, the exact observation mean itself
has primary nMSLE $2.78377\times10^{-4}$. The final EIG primary errors
are near this value despite much lower observation error. This explains
a target-dependent plateau, not a lower bound: omitting nuisance effects
can bring a prediction closer to the primary law.

\paragraph{Supplying the form versus fitting its constants.}
Table~\ref{tab:oracle-controls} uses 27 frozen 63-observation histories
per difficulty (nine prompt/design conditions and three seeds).
These earlier histories used a \texttt{gpt-4o} proposer rather than Opus;
no proposer is called during the supplied-form fits. The supplied
log-mean is
\[
\begin{aligned}
b_0+\log(\mathrm{Enz})+\log C_A+\log C_B\\
{}-\log(b_1C_B+b_2C_A+C_AC_B),
\end{aligned}
\]
where $C_A,C_B$ are substrates and $\mathrm{Enz}$ is enzyme input.
The free condition estimates $b_1,b_2$; the fixed condition supplies
them but still fits $b_0$ and noise.
\begin{table*}[tp]
\centering
\small
\setlength{\tabcolsep}{5pt}
\caption{Previously completed supplied-primary-form controls on frozen c5 histories, evaluated on 1,000 inputs. All entries are primary nMSLE.}
\label{tab:oracle-controls}
\begin{tabular}{llrrrr}
\toprule
Difficulty & Control & Histories & Minimum & Median & Maximum \\
\midrule
Easy & Constants estimated & 27 & $7.20\!\times\!10^{-7}$ & $1.20\!\times\!10^{-5}$ & $6.02\!\times\!10^{-5}$ \\
Easy & Constants supplied & 27 & $1.22\!\times\!10^{-9}$ & $8.21\!\times\!10^{-6}$ & $4.08\!\times\!10^{-5}$ \\
Hard & Constants estimated & 27 & $9.19\!\times\!10^{-3}$ & $3.11\!\times\!10^{-2}$ & $6.82\!\times\!10^{-2}$ \\
Hard & Constants supplied & 27 & $3.85\!\times\!10^{-7}$ & $9.54\!\times\!10^{-6}$ & $4.44\!\times\!10^{-5}$ \\
\bottomrule
\end{tabular}
\par\smallskip
\begin{minipage}{0.97\linewidth}\footnotesize
The same 27 saved histories per difficulty are used in each free/fixed pair.
The scale/intercept and observation noise remain fitted in the
supplied-constants condition. These older controls are not current-cohort
measurements; they are reported separately from the main experiments.
\end{minipage}
\end{table*}

In the hard case, median error is $3.11\times10^{-2}$ with free
constants versus $9.54\times10^{-6}$ with supplied constants. For the
constant whose truth is 50, posterior medians range 10.70--27.61 and all
27 reported 5th--95th percentile intervals exclude 50. This concerns the combined prior, parameterization,
data, and inference procedure, not SMC alone.

\paragraph{Prediction versus calibration.}
On ping-pong-easy (seed 31), EIG and Random have held-out negative log
likelihoods (NLLs) of $-5.173$ and $-5.167$, 90\% interval coverages of
0.771 and 0.901, and mean widths of 0.03917 and 0.05506, respectively.
Across three seeds, EIG-minus-Random NLL differences are $-0.00546$,
$-0.00765$, and $-0.00235$ nats per observation. These point-error and
log-score gains do not establish calibrated uncertainty across worlds.

\FloatBarrier

\end{document}